\documentclass[journal]{IEEEtran}
\usepackage{color,array}
\usepackage{graphicx}
\usepackage{subfigure}
\usepackage{amssymb,amsmath}
\usepackage{multirow}
\usepackage{booktabs}
\usepackage{stfloats}
\usepackage{footmisc}
\usepackage{threeparttable}
\usepackage{CJKutf8}
 
\AtBeginDocument{\begin{CJK}{UTF8}{gbsn}}
\AtEndDocument{\end{CJK}}
\newcommand{\bl}{\color{black}}

\ifCLASSINFOpdf
\else
\fi
\begin{document}
%
\title{Recognition of Handwritten Chinese Text by Segmentation: A Segment-annotation-free Approach}
%
%
%

\author{Dezhi~Peng,~
		Lianwen~Jin,~
		Weihong~Ma,~
		Canyu~Xie,~
		Hesuo~Zhang,~
		Shenggao~Zhu,~
		and~Jing~Li
\thanks{Corresponding author: Lianwen Jin.}%
\thanks{Dezhi Peng, Weihong Ma, Canyu Xie, and Hesuo Zhang are with the School of Electronics and Information Engineering, South China University of Technology, Guangzhou, Guangdong 510641, China (e-mail: pengdzscut@foxmail.com; eeweihong\_ma@mail.scut.edu.cn; eecanyuxie@mail.scut.edu.cn; eehesuo.zhang@mail.scut.edu.cn).}%
\thanks{Lianwen Jin is with the School of Electronics and Information Engineering, South China University of Technology, Guangzhou, Guangdong 510641, China and also with Peng Cheng Laboratory, Shenzhen, Guangdong, China (e-mail: lianwen.jin@gmail.com).}%
\thanks{Shenggao Zhu and Jing Li are with the Huawei Cloud Computing Technologies Company, Ltd., Shenzhen, Guangdong 518129, China (e-mail: zhushenggao@huawei.com; lijing260@huawei.com).}}

\maketitle

\begin{abstract}
Online and offline handwritten Chinese text recognition (HTCR) has been studied for decades. Early methods adopted oversegmentation-based strategies but suffered from low speed, insufficient accuracy, and high cost of character segmentation annotations. Recently, segmentation-free methods based on connectionist temporal classification (CTC) and attention mechanism, have dominated the field of HCTR. However, people actually read text character by character, especially for ideograms such as Chinese. This raises the question: are segmentation-free strategies really the best solution to HCTR? 
To explore this issue, we propose a new segmentation-based method for recognizing handwritten Chinese text that is implemented using a simple yet efficient fully convolutional network. A novel weakly supervised learning method is proposed to enable the network to be trained using only {\bl transcript} annotations; thus, the expensive character segmentation annotations required by previous segmentation-based methods can be avoided. Owing to the lack of context modeling in fully convolutional networks, we propose a contextual regularization method to integrate contextual information into the network {\bl during the training stage, which can further improve the recognition performance}. Extensive experiments conducted on four widely used benchmarks, namely CASIA-HWDB, CASIA-OLHWDB, ICDAR2013, and SCUT-HCCDoc, show that our method significantly surpasses existing methods on both online and offline HCTR, and exhibits a considerably higher inference speed than CTC/attention-based approaches.
\end{abstract}


%
\IEEEpeerreviewmaketitle

\section{Introduction}
\label{sec_intro}

\IEEEPARstart{H}{andwritten} Chinese text recognition (HCTR), including online and offline HCTR, is a challenging research topic that has been intensively studied for decades. However, owing to the large vocabulary (tens of thousands of character categories), diverse writing styles, and the character-touching problem, satisfactory recognition performance has not yet been achieved.

\begin{figure}
	\centering 
	\includegraphics[width=\columnwidth]{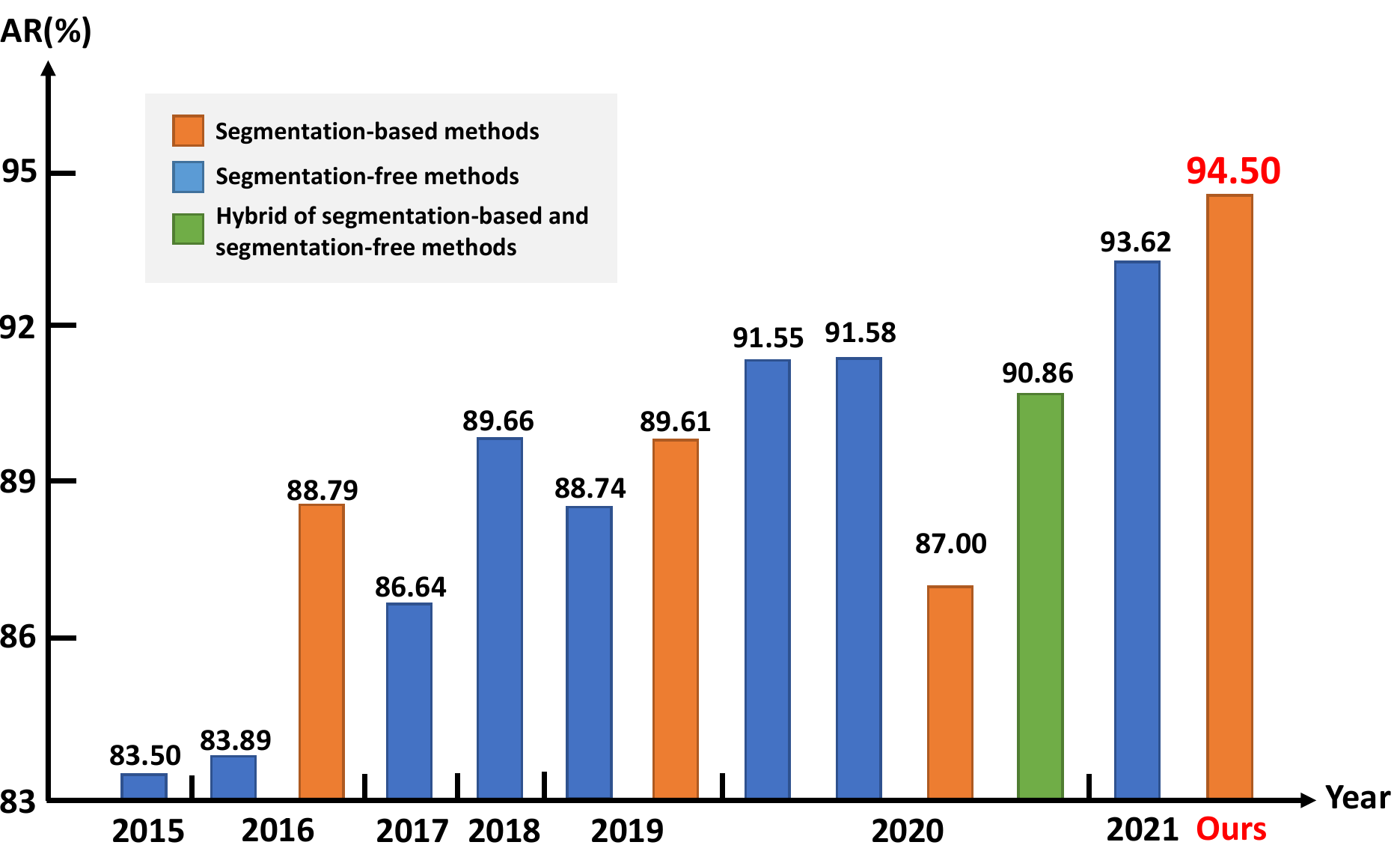}
	\caption{Performances of typical methods on the offline subset of ICDAR2013 competition dataset. Our method re-explores the segmentation-based pipeline and makes a significant improvement.} 
	\label{Fig_Intro_Methods}
\end{figure}

\begin{figure}[t]
	\centering
	\subfigure[English text]{
		\begin{minipage}[t]{1\columnwidth}
			\centering
			\includegraphics[width=\columnwidth]{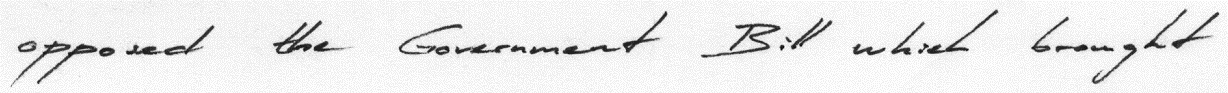}
		\end{minipage}
	}
	\subfigure[Chinese text]{
		\begin{minipage}[t]{1\columnwidth}
			\centering
			\includegraphics[width=\columnwidth]{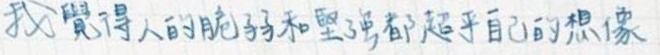}
		\end{minipage}
	}
	\caption{Example images from IAM (English) and SCUT-HCCDoc (Chinese) datasets.}
	\label{Fig_Intro_Compare}
\end{figure}

Early methods of HCTR are based on oversegmentation \cite{Q_Wang_Handwritten,X_Zhou_Handwritten,D_Wang_Approach,F_Yin_ICDAR13,X_Zhou_Minimum,S_Wang_Deep,Y_Wu_Improving,Z_Wang_Weakly}, which first oversegment the text line and then search for the best segmentation-recognition path by integrating classifier outputs, geometric context, and linguistic context. Oversegmentation-based methods used to be the most successful approaches for HCTR; however, they can be easily affected by touching or overlapping characters and require annotations to provide the boundary of characters. Recently, segmentation-free methods \cite{T_Su_Off_line,J_Du_Deep,Z_Wang_A_Comprehensive,Z_Wang_Writer,A_Graves_A_Novel,B_Shi_CRNN,R_Messina_Segmentation,Z_Xie_Fully,Y_Wu_Handwritten,K_Chen_Compact,M_Liu_Distilling,Z_Xie_Aggregation,C_Xie_High,Z_Zhu_Attention,J_Zhang_Track,J_Zhang_SRD} based on hidden Markov model (HMM), connectionist temporal classification (CTC) \cite{A_Graves_Connectionist}, and attention mechanism \cite{bahdanau2015neural}, have dominated the field of HCTR. They surpass oversegmentation-based methods and only require {\bl transcripts} to be annotated. We further illustrate the performance of typical methods on the widely used offline subset of the ICDAR2013 competition dataset \cite{F_Yin_ICDAR13} in Fig. \ref{Fig_Intro_Methods}. Although the segmentation-based method achieved considerable improvement before 2016, segmentation-free methods have become mainstream.
However, are segmentation-free approaches really the best solutions to the HCTR problem? Fig. \ref{Fig_Intro_Compare} shows two example images of handwritten English text from the IAM dataset \cite{U_Marti_IAM} and handwritten Chinese text from the SCUT-HCCDoc dataset \cite{H_Zhang_SCUT-HCCDoc}. In contrast to English texts, the basic elements of Chinese texts are characters rather than words. Intuitively, for a native Chinese speaker, each character is first segmented from the text line and then recognized when reading Chinese texts. In addition, compared with English characters, Chinese characters have a large vocabulary, diverse writing styles, complicated two-dimensional structures, and serious imbalance of character frequency, which can easily cause misalignment in segmentation-free methods, especially attention-based approaches \cite{Z_Cheng_Focusing,T_Wang_Decoupled}.
Furthermore, character segmentation information can help many downstream tasks, such as text removal, text editing, and visual information extraction. However, segmentation-free methods cannot explicitly produce character segmentation results, which limits their potential for many applications.

\begin{figure}[t]
	\centering 
	\subfigure[\bl Character segmentation (52s)]{
		\includegraphics[width=0.85\columnwidth]{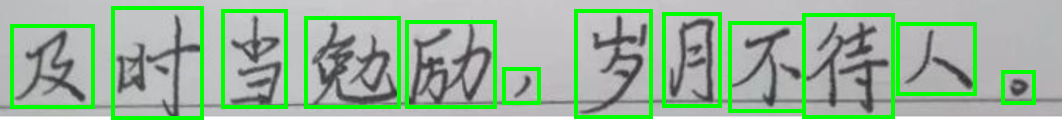}
	}
	\subfigure[\bl Transcript (5s)]{
		\includegraphics[width=0.85\columnwidth]{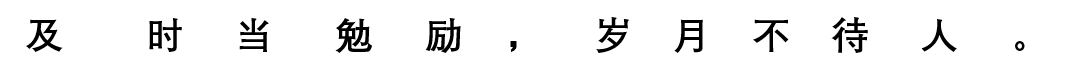}
	}
	\caption{\bl Different annotations and their time cost measured by the LabelMe\protect\footnotemark[1] \ tool.}
	\label{fig_intro_anno}
\end{figure}

In this paper, following our previous work \cite{D_Peng_Fast}, we explore a new method for recognizing handwritten Chinese text by segmentation. 
{\em First, we formulate a new segmentation-based text recognition framework for HCTR which end-to-end segments and recognizes characters through a fully convolutional network.} Compared with previous oversegmentation-based methods, the proposed framework is end-to-end trainable with high efficiency. Moreover, compared with most existing segmentation-free methods that utilize recurrent neural networks (RNNs) or auto-regressive models, our method runs in parallel and exhibits a higher inference speed. 
{\em Second, we propose a weakly supervised learning method to enable the network to be trained using only {\bl transcript} annotations.} {\bl Most existing segmentation-based methods \cite{Q_Wang_Handwritten,X_Zhou_Handwritten,D_Wang_Approach,F_Yin_ICDAR13,X_Zhou_Minimum,S_Wang_Deep,Y_Wu_Improving,D_Peng_Fast} require expensive character segmentation annotations, i.e., character bounding boxes, which are evidently much more tedious and time-consuming than transcript annotations.}
However, our method avoids such costly annotations and can still output character segmentation results. {\bl The illustration and time cost of different annotations are shown in Fig. \ref{fig_intro_anno}. }
{\em Third, a new contextual regularization method is proposed to integrate contextual information by guiding the feature extraction of the network.} 
The performance can be further boosted through the proposed contextual regularization.

Compared with the conference version \cite{D_Peng_Fast}, the major extension of this paper lies in the weakly supervised method and contextual regularization, which significantly reduce the cost of manual annotation and improve the recognition performance, respectively. 
Moreover, the conference version \cite{D_Peng_Fast} only focused on offline HCTR, whereas we further verified the effectiveness of our method on online HCTR and camera-captured handwritten text recognition.

The experiments are conducted using CASIA-HWDB \cite{C_Liu_CASIA}, CASIA-OLHWDB \cite{C_Liu_CASIA}, ICDAR2013 \cite{F_Yin_ICDAR13}, and SCUT-HCCDoc \cite{H_Zhang_SCUT-HCCDoc}. 
Our method achieves state-of-the-art performance on these datasets, which demonstrates the success of our method on both online and offline HCTR. 
Moreover, additional experiments conducted using ReCTS-25k \cite{R_Zhang_ReCTS} demonstrate the potential of our approach for scene text recognition. We hope that this paper can inspire more research to re-explore segmentation-based methods in addition to CTC and attention mechanism.

To summarize, the main contributions of this paper are as follows:
\begin{itemize}
	\item We formulate a new segmentation-based text recognition framework for online and offline HCTR. {\bl During inference,} the framework is implemented using a fully convolutional network, and thus exhibits high efficiency.
	\item We propose a weakly supervised learning method to enable the network to be trained using only {\bl transcript} annotations, thereby greatly reducing the cost of manual annotations. However, character segmentation results can still be produced using our method.
	\item We design a contextual regularization method to integrate contextual information {\bl into the training of the fully convolutional network} without slowing down the inference speed.
	\item Extensive experiments demonstrate the state-of-the-art performance of our method on both online and offline HCTR, in terms of both recognition accuracy and inference speed.
\end{itemize}

\footnotetext[1]{https://github.com/wkentaro/labelme}

\begin{figure*}[htb]
	\centering 
	\includegraphics[width=2.0\columnwidth]{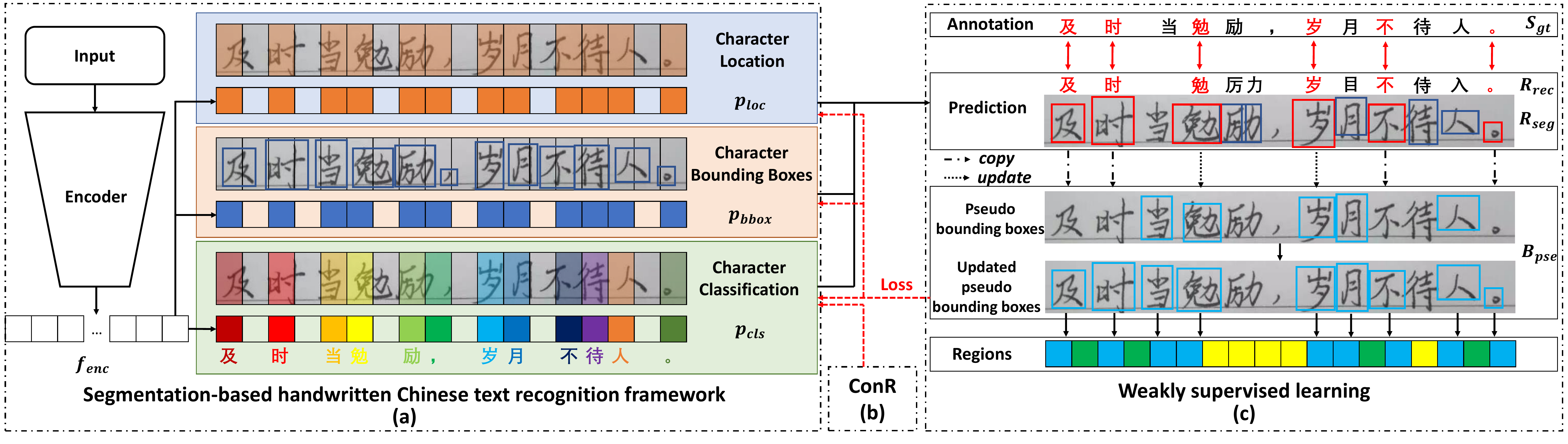}
	\caption{Proposed method consists of three parts: (a) a new segmentation-based handwritten Chinese text recognition framework, (b) a contextual regularization (ConR) term for integrating contextual information, and (c) a weakly supervised learning method for training the model using only {\bl transcript} annotations. The red dashed lines indicate the optimization process at training stage.}
	\label{Fig_Method_Overall}
\end{figure*}

\section{Related Work}
\label{sec_related_work}

\subsection{Offline Handwritten Chinese Text Recognition}
\label{sec_related_work_offline}
Offline HCTR aims to transcribe text-line images into Chinese texts. 
In general, existing methods can be categorized into segmentation-based and segmentation-free approaches.

Most previous segmentation-based (also called explicit-segmentation) approaches adopt oversegmentation-based strategies. Specifically, these methods \cite{Q_Wang_Handwritten, S_Wang_Deep} first obtain primitive segments through oversegmentation and then search for the best segmentation-recognition path. Although the character classifier, language model, and geometric model are integrated for path evaluation, satisfactory performance cannot be achieved, especially for touching or overlapping characters. Therefore, Wu et al. \cite{Y_Wu_Improving} explored neural network language models, yielding a significant improvement in the recognition performance. Furthermore, Wang et al. \cite{Z_Wang_Weakly} proposed a weakly supervised method for string-level training of oversegmentation-based systems. In addition to the strategy using oversegmentation, Peng et al. \cite{D_Peng_Fast} designed a fully convolutional network for end-to-end handwritten Chinese text segmentation and recognition.

Although enormous progress has been achieved by previous segmentation-based methods, most of them require expensive character segmentation annotations.
To address this issue, segmentation-free (also called implicit-segmentation) approaches have recently received considerable interest. One category of segmentation-free methods \cite{T_Su_Off_line,J_Du_Deep,Z_Wang_A_Comprehensive,Z_Wang_Writer} addresses the offline HCTR problem using hidden Markov model (HMM), where the handwritten text lines are modeled by a series of cascading HMMs based on the feature sequence extracted in a sliding-window manner. The method based on CTC \cite{A_Graves_Connectionist} is another popular category of solutions to offline HCTR. Messina et al. \cite{R_Messina_Segmentation} solved the offline HCTR problem by combining multi-dimensional long short-term memory (MDLSTM) and CTC. Wu et al. \cite{Y_Wu_Handwritten} further proposed separable MDLSTM to improve efficiency. Xie et al. \cite{C_Xie_High} improved the CTC-based methods by exploring data augmentation and preprocessing. {\bl Liu et al. \cite{liu2021searching} explored residual and squeeze-and-excitation structures for feature extraction and proposed context beam search to integrate the Transformer-based \cite{A_Vaswani_Attention} language model into CTC-based methods.} The attention mechanism, which has been widely adopted in scene text recognition \cite{B_Shi_Aster,T_Wang_Decoupled,luo2019moran}, action recognition \cite{H_Wu_Convolutional,J_Li_Spatio}, and video processing \cite{Na_Z_VideoWhisper}, can also be applied to offline HCTR. Xiu et al. \cite{Y_Xiu_A_Handwritten} improved the attention-based decoder by a multi-level multi-modal fusion network. In addition to the above methods that adopt a single strategy, Zhu et al. \cite{Z_Zhu_Attention} proposed a convolutional combination strategy to combine the segmentation-based and segmentation-free approaches for better performance.

\subsection{Online Handwritten Chinese Text Recognition}
\label{sec_related_work_online}

Online HCTR is aimed at recognizing Chinese text from pen-tip trajectories. In contrast to offline HCTR, the input of online HCTR contains sequential information and does not have background noise. However, the acquisition of pen-tip trajectories requires specific hardware. Owing to the rise of pen-based devices, online HCTR is also a very important research topic with wide applications in many fields such as finance and education.

The oversegmentation-based strategy can also be used to address online HCTR \cite{X_Zhou_Handwritten,D_Wang_Approach,X_Zhou_Minimum}, following the same pipeline as for offline HCTR. However, these methods rely heavily on the evaluation of candidate segmentation-recognition paths and have difficulty in recognizing touching or overlapping characters. Therefore, some studies \cite{A_Graves_A_Novel,X_Zhang_Drawing,M_Liu_Distilling} proposed to directly handle the pen-tip trajectory using RNNs and CTC. These methods extract features using long short-term memory (LSTM) or gated recurrent unit (GRU) and optimize the model with CTC loss. Liu et al. \cite{M_Liu_Distilling} further proposed distilling GRU to accelerate model training and handle handwritten texts with various styles. {\bl Instead of using RNNs, Peng et al. \cite{D_Peng_Towards} designed a global and local relation network (GLRNet) that uses self-attention \cite{A_Vaswani_Attention} for feature extraction and jointly trained the combination of GLRNet and a Transformer-based language model to achieve optimal overall performance.}
With the prevalence of convolutional neural networks (CNNs), there exist methods \cite{Z_Xie_Learning,Z_Xie_Fully,K_Chen_Compact} that transform the online pen-tip trajectory into offline feature maps that CNNs can process. Such methods first obtain image-like representations through path signature or eight-directional feature maps. Thereafter, an integrated CNN-LSTM network is trained using CTC loss. Xie et al. \cite{Z_Xie_Learning} further improved this pipeline using an implicit language model and multi-spatial context.

\section{Methodology}
\label{sec_methodology}

\subsection{Overview}
\label{sec_methodology_overview}
The overall design of the proposed method is illustrated in Fig. \ref{Fig_Method_Overall}. First, a new segmentation-based handwritten Chinese text recognition framework that segments and recognizes characters is formulated in an end-to-end manner and is implemented using an efficient fully convolutional network {\bl during inference}. Second, a novel weakly supervised learning method is proposed to effectively train the model using only {\bl transcript} annotations, which avoids costly manual segmentation annotations. Nevertheless, our model can still accurately output character segmentation results. Third, as the fully convolutional architecture can not capture contextual dependencies, a new contextual regularization (ConR) is proposed to integrate contextual information into the model {\bl during the training stage}. The proposed ConR can improve the performance without reducing the inference speed. 

\subsection{Segmentation-based Text Recognition}
\label{sec_seg_text_recog}

The proposed segmentation-based text recognition framework is illustrated in Fig. \ref{Fig_Method_Overall}(a). 
The input is text-line images for offline HCTR or image-like representations generated from pen-tip trajectories for online HCTR. For simplicity and clarity, we take offline HCTR as an example in Fig. \ref{Fig_Method_Overall}

Given an input $I_{in} \in \mathbb{R}^{H \times W \times C}$ ($H$, $W$, and $C$ denote the height, width, and number of channels of the input, respectively), the encoder extracts the feature map $f_{enc}$ as 
\begin{equation}
	\small
	f_{enc} \in \mathbb{R}^{1 \times w_{enc} \times c_{enc}} = Encoder(I_{in}),
\end{equation}
where $c_{enc}$ and $w_{enc}$ are the number of channels and the width of the feature map, respectively. The height of the feature map $f_{enc}$ is downsampled to one, and the $Encoder$ is a CNN.

Then, three predictions are produced based on the feature map $f_{enc}$ as 
\begin{equation}
	\small
	p_{loc} \in \mathbb{R}^{w_{enc} \times 1}, p_{bbox} \in \mathbb{R}^{w_{enc} \times 4}, p_{cls} \in \mathbb{R}^{w_{enc} \times n_{cls}},
\end{equation} 
where $p_{loc}$, $p_{bbox}$, and $p_{cls}$ are the character location, character bounding boxes, and character classification, respectively. The $n_{cls}$ is the total number of character categories. 
Based on these predictions, input $I_{in}$ is equally divided into $w_{enc}$ regions. The $p_{loc}^{n}$ is the confidence that the $n$-th region contains characters, $p_{bbox}^{n}$ is the coordinates of the bounding box of the character in the $n$-th region, and $p_{cls}^{n}$ contains the probabilities that the character in the $n$-th region is classified as each of the $n_{cls}$ categories. 

\begin{figure}[t]
	\centering 
	\includegraphics[width=\columnwidth]{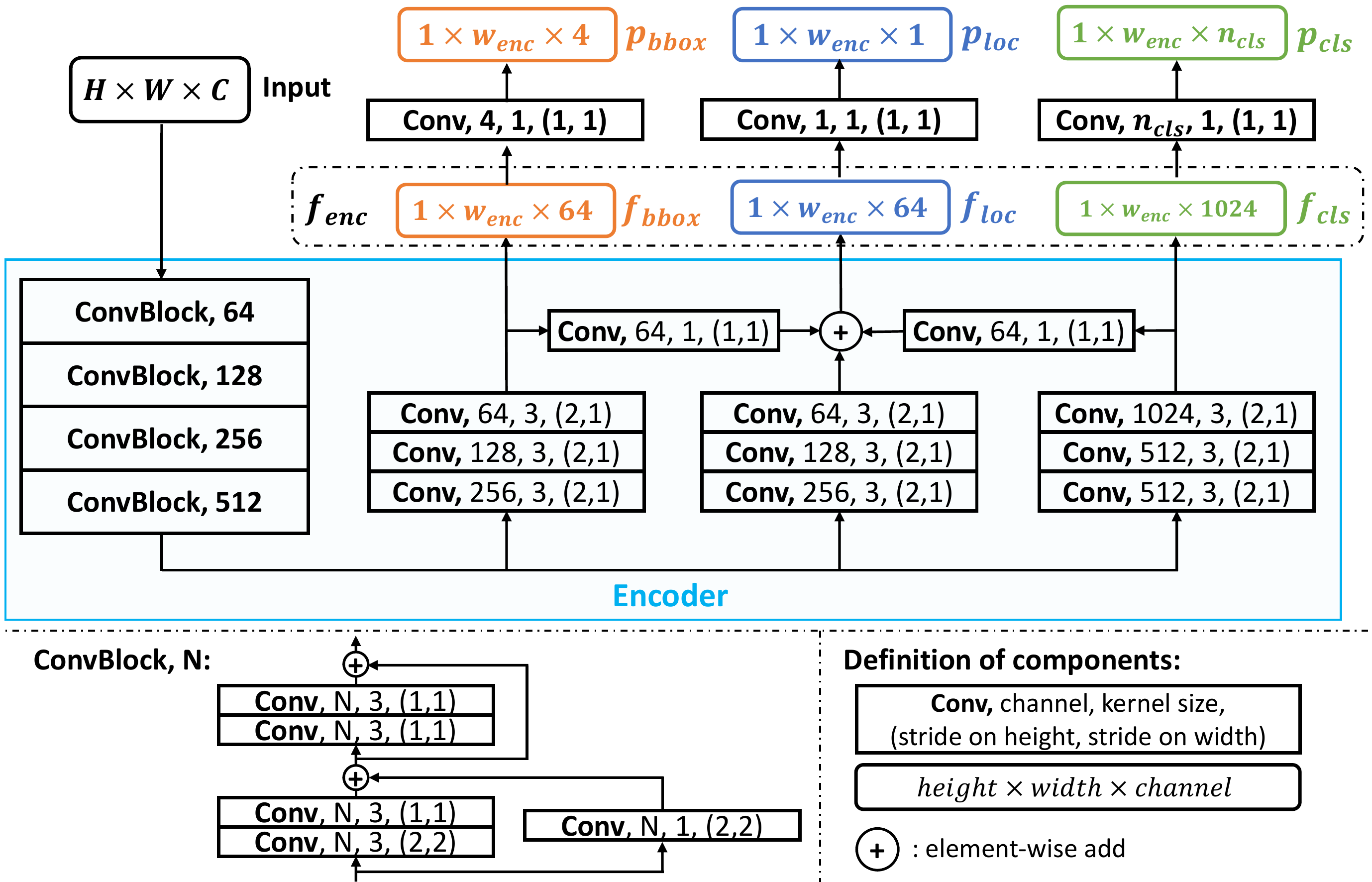}
	\caption{Detailed architecture of our network.}
	\label{Fig_Method_Arch}
\end{figure}

We implement this framework using a fully convolutional network {\bl during inference}. The detailed network architecture is illustrated in Fig. \ref{Fig_Method_Arch}. Inspired by ResNet \cite{K_He_ResNet}, the residual connection is adopted in the {\em ConvBlock}. Instead of extracting a single feature map $f_{enc}$, the encoder outputs three feature maps, $f_{loc}$, $f_{bbox}$, and $f_{cls}$, to predict the character location, character bounding boxes, and character classification, respectively.

During inference, after removing redundant predictions using non-maximum suppression (NMS) \cite{A_Neubeck_NMS} and sorting the remaining predictions from left to right, we can obtain segmentation and recognition results as shown in the ``Prediction'' part of Fig. \ref{Fig_Method_Overall}(c). Specifically, the score of the character in the $n$-th region during NMS is the weighted sum of $p_{loc}^n$ and the maximum probability in $p_{cls}^n$, in order to integrate semantic information into the character location. As specified in the conference version \cite{D_Peng_Fast}, the weight of $p_{loc}^n$ is set to 0.8.

\subsection{Weakly Supervised Learning}
\label{sec_methodology_wsl}
In this section, we propose a weakly supervised learning method (Fig. \ref{Fig_Method_Overall}(c)) to enable our network to be trained using only {\bl transcript} annotations; nevertheless, our method can still output character segmentation results that cannot be produced by segmentation-free methods.

Although character segmentation annotations are costly to annotate, it is easy to obtain isolated Chinese character samples from font files. Using these cost-free isolated character samples, we can easily synthesize text lines with character segmentation annotations. However, simply training the network using synthetic data cannot achieve satisfactory performance on real text lines, especially in complex scenarios. One solution is to develop advanced synthesis methods \cite{M_Jaderberg_Synthetic, A_Gupta_Synthetic, F_Zhan_Verisimilar, S_Fogel_ScrabbleGAN} to better mitigate the real text lines. However, for different scenarios, different data should be synthesized, and even different methods should be designed. Moreover, most existing synthesis methods produce only {\bl transcript} annotations.
Therefore, in our weakly supervised learning method, we proposed to exploit useful information from simple synthetic data to help the model learn from real data under the guidance of {\bl transcripts}.

Specifically, the network is first pretrained using synthetic data with segmentation annotations. {\bl As described in Section \ref{sec_data_synthesis},} the samples for different scenarios are synthesized in the same manner, by placing isolated characters on a white background without any complicated synthesis techniques. Although the synthetic data may differ significantly from the text lines from real scenes, the model can still learn the fundamental ability to localize and recognize characters.

Then, the real data with only {\bl transcript} annotations is also used to train the model. The procedure of the proposed weakly supervised learning method for real samples is shown in Fig. \ref{Fig_Method_Overall}(c). Similar to the learning process of humans, the model is taught which prediction is correct and trained based on past successful experiences.
Specifically, for a real input, the network predicts the segmentation result $R_{seg}$ and recognition result $R_{rec}$ as
\begin{equation}
	\small
	R_{seg} = \{(r_{seg}^{1}, r_{sco}^{1}), (r_{seg}^{2}, r_{sco}^{2}), ..., (r_{seg}^{l_{pr}}, r_{sco}^{l_{pr}})\},
\end{equation}
\vspace{-1em}
\begin{equation}
	\small
	R_{rec} = \{r_{rec}^{1}, r_{rec}^{2}, ..., r_{rec}^{l_{pr}}\},
\end{equation}
where $r_{seg}^{i}$, $r_{sco}^{i}$, and $r_{rec}^{i}$ are the coordinates of the bounding box, score, and category of the $i$-th predicted character, respectively, and $l_{pr}$ is the total number of predicted characters.

Thereafter, we compute the edit distance between {\bl transcript} annotation $S_{gt}$ and recognition result $R_{rec}$, where $S_{gt}$ is defined as 
\begin{equation}
	\small
	S_{gt} = \{s_{gt}^{1}, s_{gt}^{2}, ..., s_{gt}^{l_{gt}}\},
\end{equation}
where $s_{gt}^j$ is the category of the $j$-th character in the {\bl transcript}, and $l_{gt}$ is the total number of annotated characters.
The characters, which are matched as ``equal'' in computing edit distance, are marked using red arrows between the annotation and prediction in Fig. \ref{Fig_Method_Overall}(c).
Because character recognition and segmentation tasks are highly related, the ``equal'' characters in the prediction are very likely to have accurate bounding box predictions.

Next, the bounding boxes corresponding to ``equal'' characters (red bounding boxes in Fig. \ref{Fig_Method_Overall}(c)) are used to update the pseudo bounding boxes. The pseudo bounding boxes $B_{pse}$ are defined as follows:
\begin{equation}
	\small
	B_{pse} = \{(b_{pse}^{1}, b_{sco}^{{1}}), (b_{pse}^{2}, b_{sco}^{{2}}), ..., (b_{pse}^{l_{gt}}, b_{sco}^{{l_{gt}}})\},
\end{equation}
where $b_{pse}^{j}$ and $b_{sco}^{{j}}$ are the coordinates and score of the pseudo bounding box of character $s_{gt}^{j}$, respectively. All the pseudo bounding boxes are initialized as $\varnothing$. If the character $r_{rec}^{i}$ is matched to be ``equal'' to the character $s_{gt}^{j}$, the segmentation result $r_{seg}^{i}$ is used to update the pseudo bounding box as follows:
\begin{equation}
	\small
	b_{pse}^{j} = \left\{
	\begin{aligned}
		& r_{seg}^{i}, & b_{pse}^{j} = \varnothing, \\
		& \lambda_{pse} \times b_{pse}^{j} + (1 - \lambda_{pse}) \times r_{seg}^{i}, & otherwise,
	\end{aligned}
	\right.
\end{equation}
\begin{equation}
	\small
	b_{sco}^{j} = \left\{
	\begin{aligned}
		& r_{sco}^{i}, & b_{pse}^{j} = \varnothing, \\
		& \lambda_{pse} \times b_{sco}^{j} + (1 - \lambda_{pse}) \times r_{sco}^{i}, & otherwise,
	\end{aligned}
	\right.
\end{equation}
where $\lambda_{pse}$ is calculated as:
\begin{equation}
	\small
	\lambda_{pse} = \frac{e^{10 \times b_{sco}^{j}}}{e^{10 \times b_{sco}^{j}} + e^{10 \times r_{sco}^{i}}}.
\end{equation}
The weight $\lambda_{pse}$ is a function of $b_{sco}^{j}$ and $r_{sco}^{i}$, so as to make the bounding box with a higher score have a much higher weight. 

Subsequently, the pseudo bounding boxes $B_{pse}$ and {\bl transcript} annotation $S_{gt}$ are used to optimize the network. However, there may exist pseudo bounding boxes equal to $\varnothing$, which implies that the loss cannot be computed in a normal manner. Therefore, a special method for loss calculation is designed.

Specifically, we first project the pseudo bounding boxes that are not equal to $\varnothing$ onto their corresponding regions of the model input, yielding a mapping $M_{ptr}$.
The element $(j, n) \in M_{ptr}$ indicates that the center point of the pseudo bounding box $b_{pse}^{j}$ is within the $n$-th region. The regions that contain the center points of pseudo bounding boxes are represented by the blue squares in the ``Regions'' part of Fig. \ref{Fig_Method_Overall}(c). Then the loss of $p_{bbox}$ and $p_{cls}$ are calculated as:
\begin{equation}
	\small
	l_{bbox} = \frac{1}{|M_{ptr}|}\sum_{(j, n) \in M_{ptr}}SE(b_{pse}^{j}, p_{bbox}^{n}),
\end{equation}
\begin{equation}
	\small
	\label{eq_l_cls}
	l_{cls} = -\frac{1}{|M_{ptr}|}\sum_{(j, n) \in M_{ptr}}log(p_{cls}^{n, s_{gt}^{j}}),
\end{equation}
where the function $SE$ calculates the square error between two inputs.

Regarding the loss of character location $p_{loc}$, we only know that the blue regions in Fig. \ref{Fig_Method_Overall}(c) contain characters. Owing to the existence of pseudo bounding boxes equal to $\varnothing$, the difficulty lies in determining the negative samples, i.e., the regions that do not contain characters. Fortunately, although we cannot find all the negative samples, it can be confirmed that there is no character in the regions between two consecutive pseudo bounding boxes. Specifically, if both $b_{pse}^{j}$ and $b_{pse}^{j+1}$ are not equal to $\varnothing$, the indices of the regions, which are between the two regions corresponding to $b_{pse}^{j}$ and $b_{pse}^{j+1}$, are added to the set $N_{loc}$. The regions in $N_{loc}$ contain no characters and are represented by the green squares in Fig. \ref{Fig_Method_Overall}(c).
However, whether the yellow regions contain characters cannot be determined because of missing pseudo bounding boxes. Thus, these regions are not considered in the loss calculation. Consequently, the loss of $p_{loc}$ is formulated as:
\begin{equation}
	\small
	l_{loc} = {\bl -\frac{0.5}{|T_{loc}|} \sum_{n \in T_{loc}}log(p_{loc}^{n})} - \frac{0.5}{|N_{loc}|} \sum_{n \in N_{loc}}log(1 - p_{loc}^{n}),
\end{equation} 
{\bl where $T_{loc}$ is the set of the indices of the blue regions that are supposed to contain characters.}

Finally, the total loss is given by:
\begin{equation}
	\small
	l_{total} = l_{bbox} + l_{cls} + l_{loc}.
\end{equation}

\subsection{Contextual Regularization}
The fully convolutional architecture results in high efficiency and parallel computing, but it is difficult to capture contextual dependencies.
The prediction for each region of a text line depends only on the corresponding receptive field. For example, 
the character classification $p_{cls}$ can be reformulated as:
\begin{equation}
	\small
	\begin{aligned}
		p_{cls} & = \{p_{cls}^{1}, ..., p_{cls}^{w_{enc}}\} = FCN(I_{in}) \\
		& = \{FCN(F^{1}), ..., FCN(F^{w_{enc}})\},
	\end{aligned}
\end{equation}
where $FCN$ is our network and $F^{n}$ is the receptive field of the $n$-th character classification prediction. In the above formula, the predictions in $p_{cls}$ are independent without considering the relationship between them.

\begin{figure}[b]
	\centering
	\includegraphics[width=1\columnwidth]{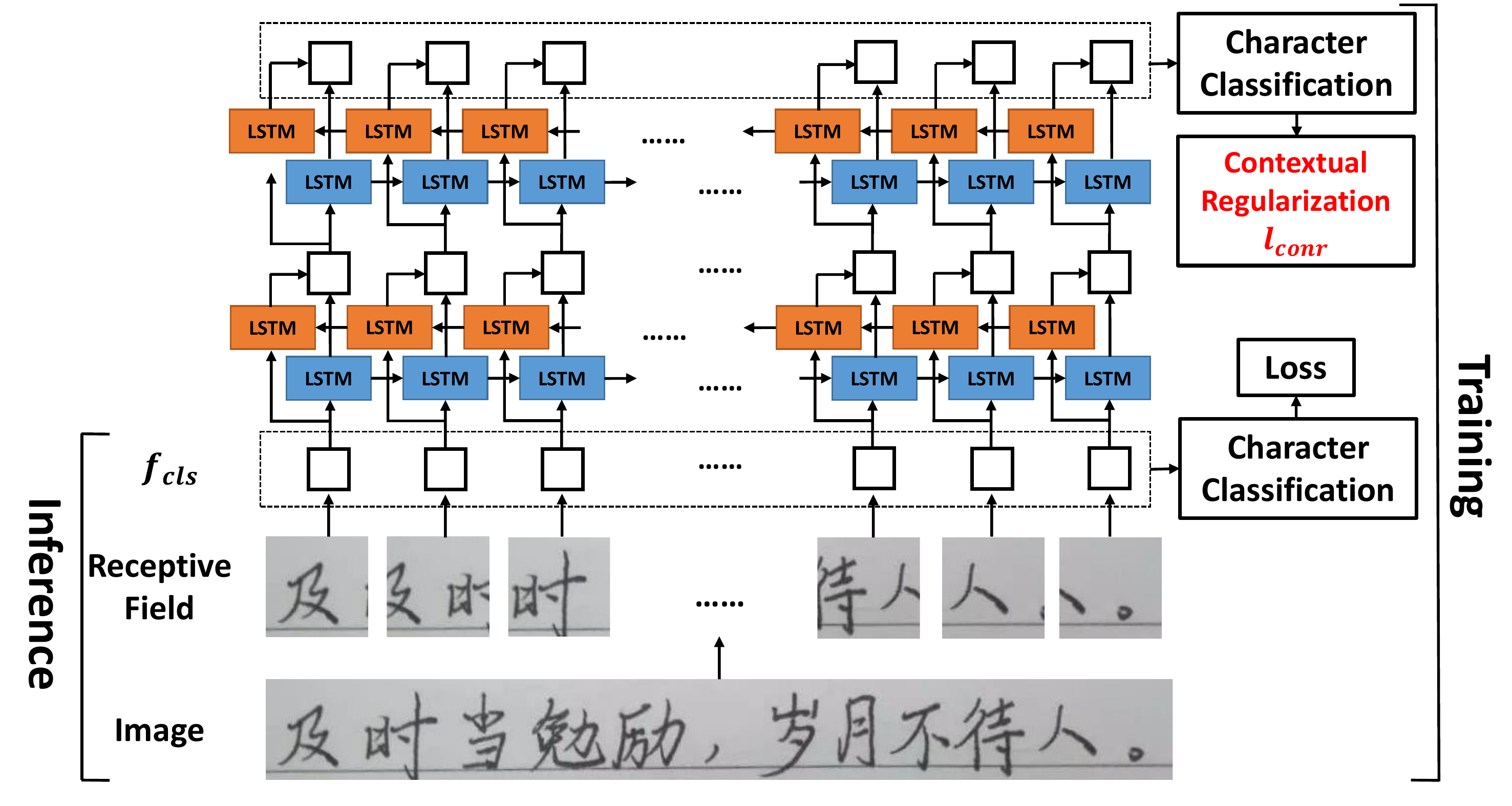}
	\caption{Illustration of the contextual regularization.}
	\label{Fig_Method_ConR}
\end{figure}

However, the task of HCTR, especially the character classification part, relies heavily on contextual information. Even for native speakers, it is difficult to recognize some confusing cursive handwritten characters without context. 
Previous methods adopted recurrent layers, such as LSTM and GRU, for context modeling.
However, these recurrent layers can not run in parallel, which greatly reduces the inference speed especially when handling long texts. 
To this end, we propose to solve context modeling from a new perspective, by guiding the feature extraction using a novel contextual regularization (ConR) {\bl only in the training stage}, as illustrated in Fig. \ref{Fig_Method_ConR}.

During training, two bidirectional long short-term memory (BLSTM) layers are added on top of the feature map $f_{cls}$. Subsequently, an extra character classification is predicted based on the output of the BLSTM layers. The loss $l_{conr}$ of this new character classification is calculated in the same way as for $l_{cls}$ (Eq. (\ref{eq_l_cls})). Then, the loss $l_{conr}$ is added to the total loss $l_{total}$ as a regularization term. Through the backpropagation of the gradient, ConR can guide the feature $f_{cls}$ to capture contextual information with the help of the context modeling ability of BLSTM layers.

During inference, the BLSTM layers and subsequent character classification are removed to maintain the high processing speed of the fully convolutional architecture. Experiments show that ConR results in a consistent improvement on all datasets. Even if the BLSTM layers and subsequent character classification are used during inference, the performance remains nearly unchanged, which proves that the feature map $f_{cls}$ really learns contextual information.

\section{Experiments}
\label{sec_experiments}

\subsection{Datasets}
\noindent\textbf{CASIA-HWDB} \cite{C_Liu_CASIA} is a large-scale offline handwriting database, including CASIA-HWDB1.0-1.2 and CASIA-HWDB2.0-2.2. CASIA-HWDB1.0-1.2 contain 3,895,135 isolated characters from 1,020 writers, while CASIA-HWDB2.0-2.2 include 52,230 text lines from 1,019 writers. Note that the isolated character samples of CASIA-HWDB1.0-1.2 are not the characters cropped from the text lines of CASIA-HWDB2.0-2.2. 

\noindent\textbf{CASIA-OLHWDB} \cite{C_Liu_CASIA} is the online version of CASIA-HWDB, which consists of CASIA-OLHWDB1.0-1.2 and CASIA-OLHWDB2.0-2.2. CASIA-OLHWDB1.0-1.2 contain 3,912,017 isolated characters from 1,020 writers while CASIA-OLHWDB2.0-2.2 contain 52,220 text lines from 1,019 writers. The CASIA-OLHWDB2.0-2.2 are further divided into 41,710 text lines for training and 10,510 text lines for testing.

\noindent\textbf{ICDAR2013} competition dataset \cite{F_Yin_ICDAR13} contains 3,432 online and offline handwritten Chinese text lines from 60 writers. For convenience, the online and offline subsets are denoted as \textbf{ICDAR2013-Online} and \textbf{ICDAR2013-Offline}, respectively.

\noindent\textbf{SCUT-HCCDoc} \cite{H_Zhang_SCUT-HCCDoc} contains 12,253 offline camera-captured document images with 116,629 text lines. The training and testing sets comprise 93,411 text lines and 23,218 text lines, respectively.

\begin{figure}[tb]
	\centering
	\subfigure[For CASIA-HWDB]{
		\begin{minipage}[t]{1\columnwidth}
			\centering
			\includegraphics[width=1\columnwidth]{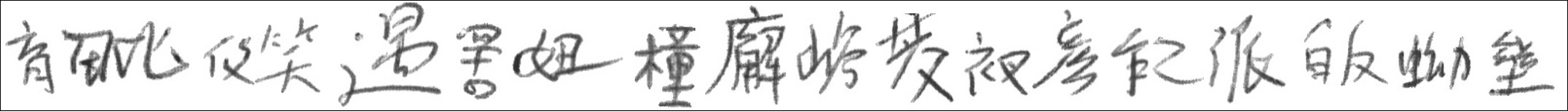}
		\end{minipage}
	}	
	\subfigure[For CASIA-OLHWDB]{
		\begin{minipage}[t]{\columnwidth}
			\centering 
			\includegraphics[width=\columnwidth]{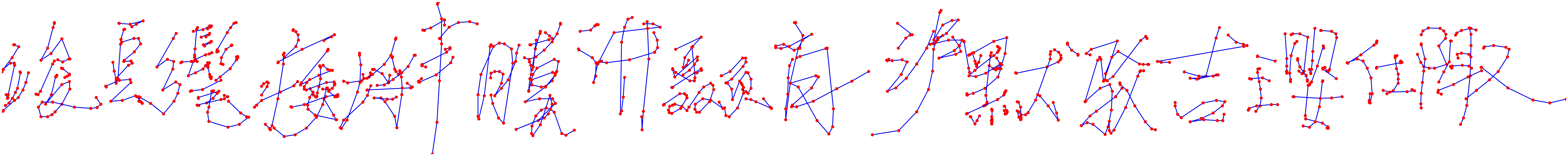}
		\end{minipage}
	}
	\subfigure[For SCUT-HCCDoc]{
		\begin{minipage}[t]{1\columnwidth}
			\centering
			\includegraphics[width=0.8\columnwidth]{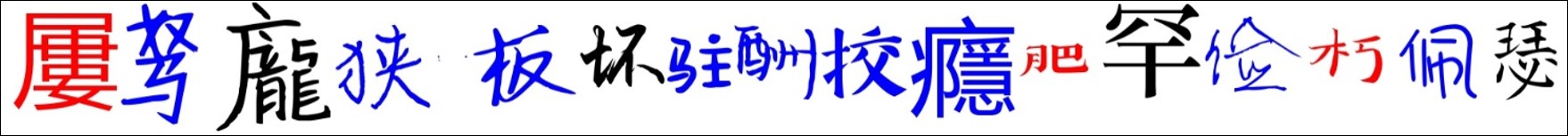}
		\end{minipage}
	}
	\caption{Examples of synthetic samples (without semantic information) for CASIA-HWDB, CASIA-OLHWDB, and SCUT-HCCDoc.}
	\label{Fig_Exp_Synth}
\end{figure}

\subsection{Data Synthesis}
\label{sec_data_synthesis}
For all real datasets, synthetic text lines are synthesized by simply placing offline characters on a white background or concatenating the pen-tip trajectories of online characters. The synthetic samples for CASIA-HWDB and CASIA-OLHWDB use isolated characters from CASIA-HWDB1.0-1.2 and CASIA-OLHWDB1.0-1.2, respectively, while the synthetic samples for SCUT-HCCDoc use the character images generated from 101 font files. 
The categories of characters are randomly selected from the vocabulary when synthesizing data without semantic information, while the corpora described in Section \ref{sec_exp_transcription} are adopted when synthesizing data with semantic information.
Fig. \ref{Fig_Exp_Synth} shows examples of synthetic samples (without semantic information) for CASIA-HWDB, CASIA-OLHWDB, and SCUT-HCCDoc.

\subsection{Implementation Details}
\label{sec_implement_detail}
We conduct our experiments using an NVIDIA GTX 1080ti GPU with 11GB of memory and implement our method using PyTorch. 
First, in the pretraining stage, the model is pretrained using synthetic data for 150,000 iterations. The weakly supervised learning method and ConR are not adopted in this stage.
Then, in the training stage, the model is trained using both real and synthetic data for 1,200,000 iterations.
The synthetic data is synthesized on the fly.
The network is optimized using stochastic gradient descent (SGD) with a batch size of 8 and an initial learning rate of 0.01. The learning rate is multiplied by 0.1 at 25\%, 50\%, and 75\% of the total number of iterations. During inference, the batch size is set to 1.

\subsection{Transcription}
\label{sec_exp_transcription}
During inference, the transcription without language model is described in the last paragraph of Section \ref{sec_seg_text_recog}, which first removes redundant predictions through NMS and then rearranges the remaining characters from left to right. In the following sections, if not specified, the results are obtained by the transcription without language model.

Moreover, the transcription process can be combined with n-gram language models. Specifically, a tri-gram language model generated from the same corpora as the conference version \cite{D_Peng_Fast} is adopted. The corpora consist of the PFR corpus \cite{PFR} (news text of 2,199,492 characters from the 1998 People's Daily corpus), the PH corpus \cite{PH} (news text of 3,697,028 characters from the People's Republic of China's Xinhua news recorded between January 1990 and March 1991), and the CLDC corpus \cite{CLDC} (50 million characters collected by the Institute of Applied Linguistics). Because the character classification prediction $p_{cls}$ contains the probability of each character category and the blank probability can be calculated as $1-p_{loc}$, the CTC beam search algorithm \cite{A_Graves_Towards} can be used for the transcription with the tri-gram language model based on the CTC-style predictions formed by $p_{cls}$ and $1-p_{loc}$. 

{\bl Recently, the Transformer-based \cite{A_Vaswani_Attention} language model has emerged in the field of HCTR \cite{Y_Xiu_A_Handwritten,D_Peng_Towards,liu2021searching}. After preparing CTC-style predictions as mentioned above, the transcription process can also be integrated with the Transformer-based language model through the context beam search algorithm \cite{liu2021searching}. The Transformer-based language model follows the architecture specified in \cite{liu2021searching} and is trained with the same corpora as the tri-gram language model using the Fairseq \cite{ott2019fairseq} toolkit.}
\subsection{Evaluation Metrics}

Following previous studies on online and offline HCTR, the accurate rate (AR) and correct rate (CR) are adopted to evaluate the performance of methods, which are calculated as
\begin{equation}
	\small
	\begin{aligned}
		& AR &= & \ (N_t - D_e - S_e - I_e) / N_t, \\
		& CR &= & \ (N_t - D_e - S_e) / N_t, 
	\end{aligned}
\end{equation}
where $D_e$, $S_e$, and $I_e$ represent the total number of deletion, substitution, and insertion errors, respectively, and $N_t$ is the total number of characters in the annotations. The errors are calculated between the recognition result and {\bl transcript} annotation. The AR and CR are presented as percentages in the tables of the following sections.

\begin{figure}[b]
	\centering 
	\includegraphics[width=\columnwidth]{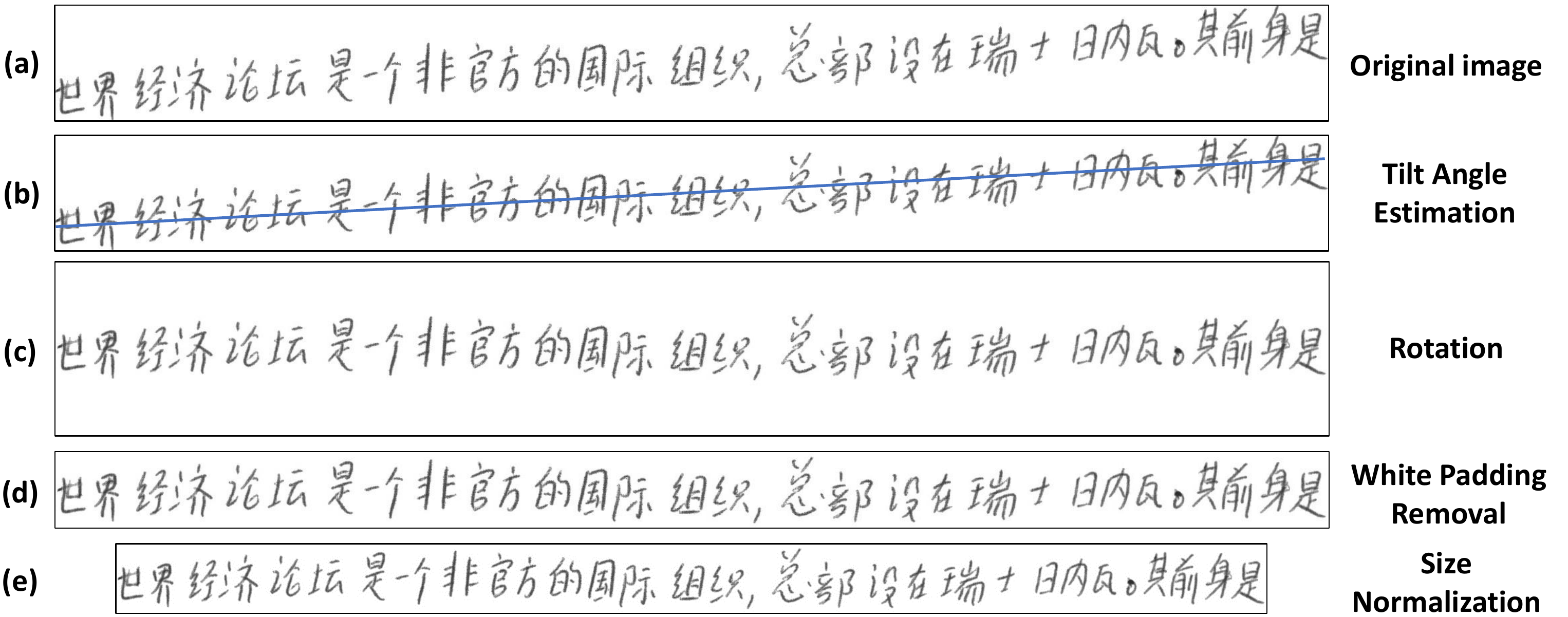}
	\caption{Illustration of data preproccessing for the text-line images from CASIA-HWDB2.0-2.2 and ICDAR2013-Offline.}
	\label{fig_exp_prepro_icdar2013-offline}
\end{figure}

\subsection{Experiments on ICDAR2013-Offline}
\label{sec_exp_icdaroffline}

\subsubsection{Experimental Settings}
The model is first pretrained using the synthetic data (without semantic information) for CASIA-HWDB. Then, the 52,230 real samples from CASIA-HWDB2.0-2.2 and the synthetic data (with semantic information) for CASIA-HWDB are used to train the model. Both the ratios of the real and synthetic samples in a batch are 0.5. Finally, the model is evaluated on the 3,432 samples from ICDAR2013-Offline. Following the setting of most previous methods, the number of character categories is set to 7,356.

\subsubsection{Data Preprocessing}
For the text-line images from CASIA-HWDB2.0-2.2 and ICDAR2013-Offline, the data preprocessing is illustrated in Fig. \ref{fig_exp_prepro_icdar2013-offline}. Given a text-line image (Fig. \ref{fig_exp_prepro_icdar2013-offline}(a)), we first estimate the tilt angle of the text by conducting linear regression using the coordinates of black pixels (Fig. \ref{fig_exp_prepro_icdar2013-offline}(b)). Then, the text-line image is rotated to make the text horizontal (Fig. \ref{fig_exp_prepro_icdar2013-offline}(c)). Next, we remove the white padding at the top and bottom of the image to highlight the text (Fig. \ref{fig_exp_prepro_icdar2013-offline}(d)). Finally, the height of the text-line image is normalized to 128 pixels while maintaining the aspect ratio (Fig. \ref{fig_exp_prepro_icdar2013-offline}(e)). 

For the synthetic text-line images, only size normalization (Fig. \ref{fig_exp_prepro_icdar2013-offline}(e)) is performed.

\subsubsection{Experimental Results}
\label{sec_icdaroffline_results}

\begin{table}[tb]
	\centering
	\caption{Comparison with existing methods on ICDAR2013-Offline (LM: language model)}
	\label{tbl_exp_ICDAR2013-offline}
	\begin{tabular}{lcccc}
		\hline 
		\multirow{2}*{Method} & \multicolumn{2}{c}{Without LM} & \multicolumn{2}{c}{With LM} \\
		\cmidrule(lr){2-3} \cmidrule(lr){4-5}
		& AR & CR & AR & CR \\ 
		\hline
		HIT-2 \cite{F_Yin_ICDAR13} & - & - & 86.73 & 88.76 \\
		Messina et al. \cite{R_Messina_Segmentation} & 83.50 & - & 89.40 & - \\
		Wu et al. \cite{Y_Wu_Handwritten} & 86.64 & 87.43 & 90.38 & - \\ 
		Du et al. \cite{J_Du_Deep} & 83.89 & - & 93.50 & - \\ 
		Wang et al. \cite{S_Wang_Deep} & 88.79 & 90.67 & 94.02 & 95.53 \\
		Wu et al. \cite{Y_Wu_Improving} & - & - & 96.20 & 96.32 \\
		Wang et al. \cite{Z_Wang_A_Comprehensive} & 89.66 & - & 96.47 & - \\
		Xie et al. \cite{Z_Xie_Aggregation} & 91.25 & 91.68 & 96.22 & 96.70 \\
		Peng et al. \cite{D_Peng_Fast} & 89.61 & 90.52 & 94.88 & 95.51 \\
		Xiu et al. \cite{Y_Xiu_A_Handwritten} & 88.74 & - & 96.35 & - \\ 
		Xie et al. \cite{C_Xie_High} & 91.55 & 92.13 & 96.72 & 96.99 \\ 
		Wang et al. \cite{Z_Wang_Writer} & 91.58 & - & 96.83 & - \\
		Wang et al. \cite{Z_Wang_Weakly} & 87.00 & 89.12 & 95.11 & 95.73 \\
		Zhu et al. \cite{Z_Zhu_Attention} & 90.86 & - & 94.00 & - \\
		{\bl Liu et al. \cite{liu2021searching}} & {\bl 93.62} & - & {\bl 97.51} & - \\  
		\hline
		\textbf{Ours (tri-gram LM)} & \textbf{94.50} & \textbf{94.76} & 96.79 & 97.32\\
		\textbf{\bl Ours (Transformer-based LM)} & \textbf{\bl 94.50} & \textbf{\bl 94.76} & \textbf{\bl 97.70} & \textbf{\bl 97.91}\\
		\hline
	\end{tabular}
\end{table}

\begin{figure}[b]
	\centering 
	\includegraphics[width=\columnwidth]{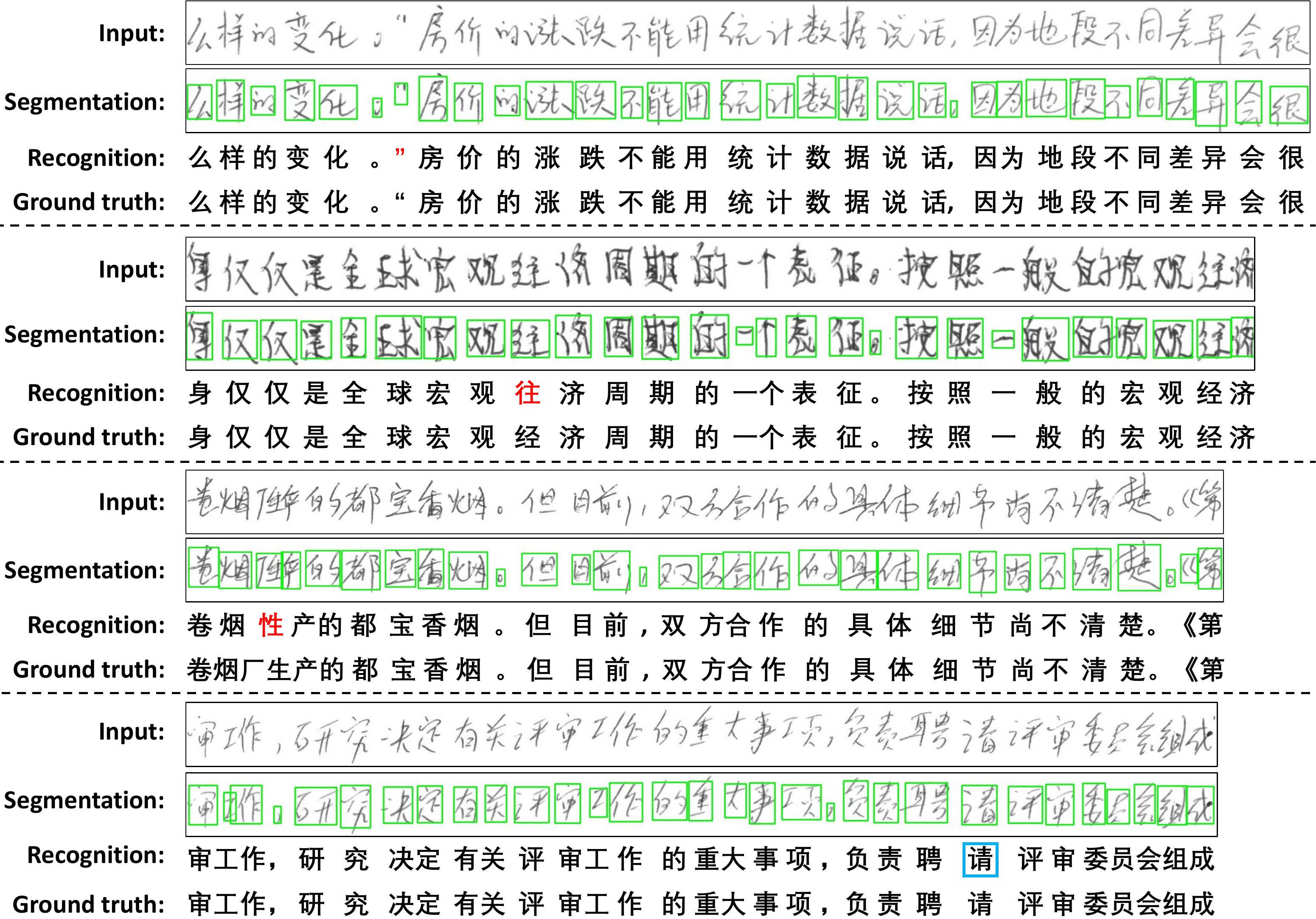}
	\caption{Visualization results of ICDAR2013-Offline. The recognition results without language model are presented.}
	\label{fig_exp_vis_ICDAR2013-offline}
\end{figure}

Table \ref{tbl_exp_ICDAR2013-offline} shows the comparison of existing methods and ours on ICDAR2013-Offline. {It can be seen that our approach achieves state-of-the-art performance with and without language model.}
{\bl Specifically, taking advantage of the Transformer-based language model, the method in \cite{liu2021searching} outperforms our approach that uses the traditional tri-gram language model. However, state-of-the-art performance can still be achieved when our method is also equipped with the Transformer-based language model.}

Moreover, in addition to the outstanding recognition performance, our method can also output character segmentation results. Fig. \ref{fig_exp_vis_ICDAR2013-offline} shows the visualization results of ICDAR2013-Offline. It can be seen that the characters can be segmented accurately from the text-line image. {\bl Particularly in the fourth example, the character ``请'' is written very similar to ``清'' but is still correctly recognized (marked by blue square), demonstrating the ability to distinguish similar characters. This ability may be attributed to the contextual information integrated by ConR and the large-scale synthetic datasets. Nonetheless, there are still some misrecognized similar characters, e.g., ``经'' is recognized as ``往'' in the second example, indicating the recognition of similar handwritten Chinese characters is still an important issue worth studying in the future.}

\subsection{Experiments on ICDAR2013-Online}
\label{sec_exp_icdar2013_online}
\subsubsection{Experimental Settings}
First, the synthetic data (without semantic information) for CASIA-OLHWDB is used to pretrain the model. Then, the model is trained using the 41,710 real samples from the training set of CASIA-OLHWDB2.0-2.2 and the synthetic data (with semantic information) for CASIA-OLHWDB. Both the ratios of the real and synthetic samples in a batch are 0.5. After training, the 3,432 samples from ICDAR2013-Online are used for evaluation. Following previous studies, the number of character categories is set to 7,356.

\begin{figure}[t]
	\centering 
	\includegraphics[width=0.9\columnwidth]{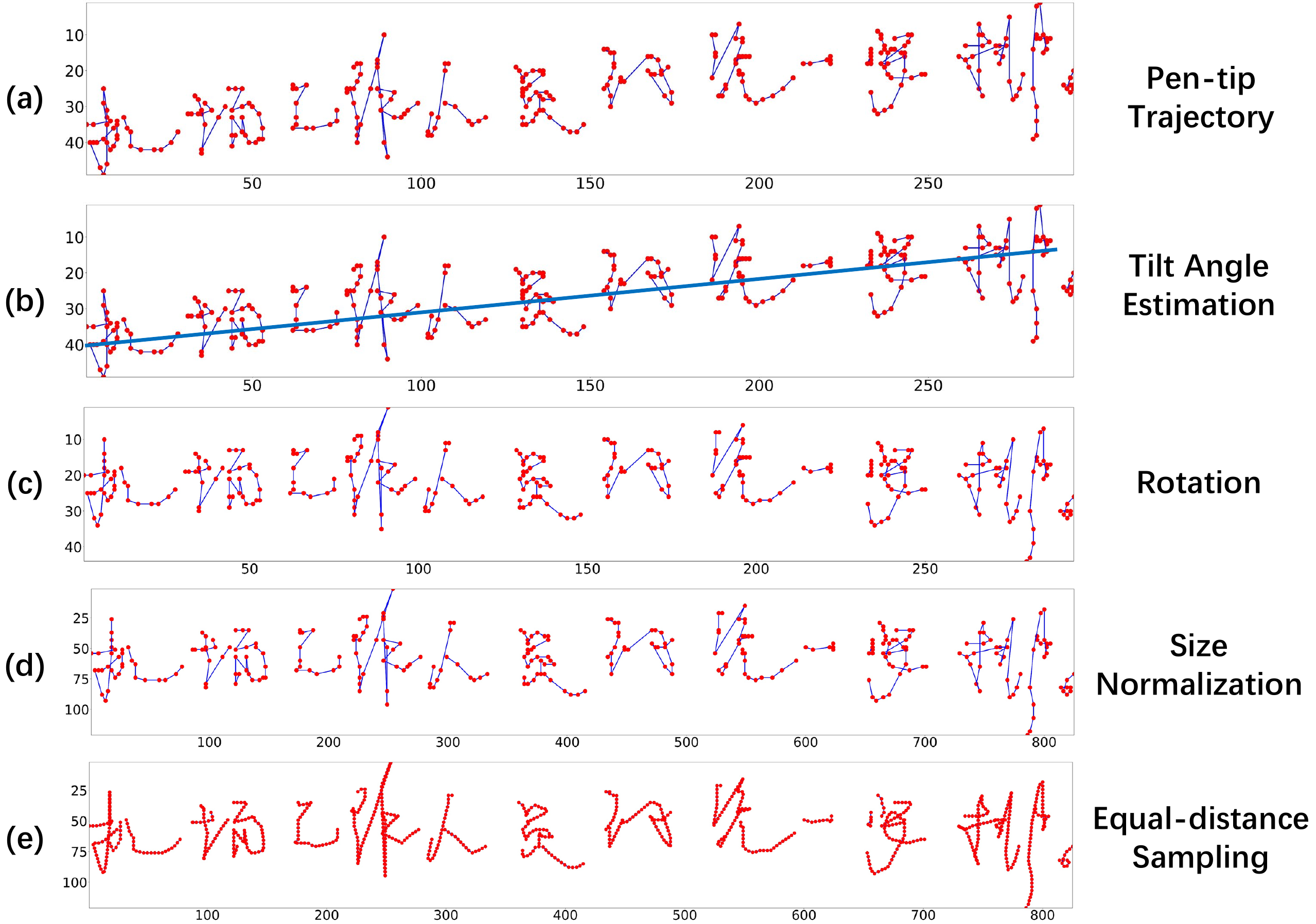}
	\caption{Illustration of data preprocessing for the pen-tip trajectories from CASIA-OLHWDB2.0-2.2 and ICDAR2013-Online.}
	\label{fig_exp_prepro_icdar2013-online}
\end{figure}

\begin{figure}[b]
	\centering
	\includegraphics[width=0.6\columnwidth]{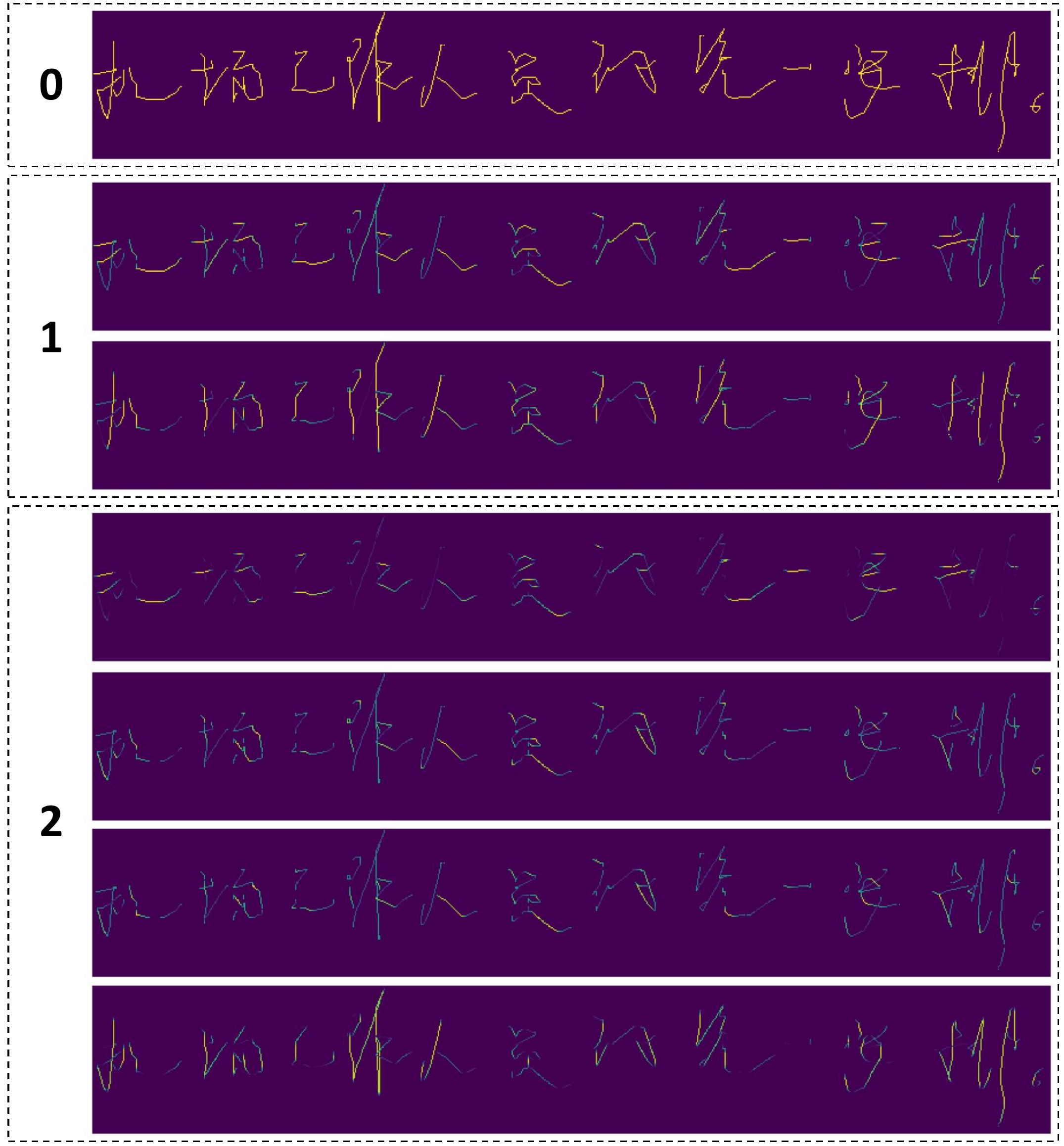}
	\caption{Visualization of path signature feature maps (up to 2nd order).}
	\label{fig_exp_vis_ps}
\end{figure}

\subsubsection{Data Preprocessing}
The pipeline of data preprocessing for pen-tip trajectories from CASIA-OLHWDB2.0-2.2 and ICDAR2013-Online is depicted in Fig. \ref{fig_exp_prepro_icdar2013-online}. Given a pen-tip trajectory containing a sequence of points belonging to multiple strokes (Fig. \ref{fig_exp_prepro_icdar2013-online}(a)), we first perform linear fitting based on the $(x, y)$ coordinates of the points and obtain the tilt angle of the text line, as shown in Fig. \ref{fig_exp_prepro_icdar2013-online}(b). Then, the text line is rotated to be horizontal, as depicted in Fig. \ref{fig_exp_prepro_icdar2013-online}(c). Next, the $(x,y)$ coordinates of the points are rescaled to normalize the height of the text line to 128 while maintaining the aspect ratios, as shown in Fig. \ref{fig_exp_prepro_icdar2013-online}(d). Because the raw data adopts uniform-time sampling, the density of the points is related to the writing speed and sampling rate. Therefore, we resample the trajectory in an equal-distance manner as shown in Fig. \ref{fig_exp_prepro_icdar2013-online}(e). Specifically, the points are sampled at intervals of a Euclidean distance of 1.

For the synthetic pen-tip trajectories, only size normalization (Fig. \ref{fig_exp_prepro_icdar2013-online}(d)) and equal-distance sampling (Fig. \ref{fig_exp_prepro_icdar2013-online}(e)) are adopted.

\subsubsection{Path Signature}
\label{sec_exp_ps}
For online HCTR, it is crucial to translate the pen-tip trajectory into offline feature maps while retaining most of the online information. The path signature \cite{K_Chen_Integration,T_Lyons_System,T_Lyons_Rough,B_Hambly_Uniqueness} has been verified to be effective in both online Chinese character recognition \cite{B_Graham_Sparse,W_Yang_Improved,W_Yang_DropSample} and online HCTR \cite{Z_Xie_Fully,K_Chen_Compact,Z_Xie_Learning}. Therefore, following previous methods \cite{Z_Xie_Learning}, the truncated path signature feature maps up to $2$nd integrated integral are calculated in a sliding-window fashion with a window size of 9. Fig. \ref{fig_exp_vis_ps} visualizes the generated path signature feature maps.

\begin{table}[t]
	\centering
	\caption{Comparison with existing methods on ICDAR2013-Online (LM: language model)}
	\label{tbl_exp_ICDAR2013-online}
	\begin{tabular}{lcccc}
		\hline
		\multirow{2}*{Method} & \multicolumn{2}{c}{Without LM} & \multicolumn{2}{c}{With LM} \\
		\cmidrule(lr){2-3} \cmidrule(lr){4-5}
		& AR & CR & AR & CR \\
		\hline 
		Zhou et al. \cite{X_Zhou_Handwritten} & - & - & 94.06 & 94.76 \\
		Zhou et al. \cite{X_Zhou_Minimum} & - & - & 94.22 & 94.76 \\
		Sun et al. \cite{L_Sun_Deep} & 89.12 & 90.18 & 93.40 & 94.43 \\
		2C-FCRN+impLM \cite{Z_Xie_Learning} & 88.88 & 90.17 & 95.46 & 96.01 \\
		2C-FCRN+impLM\&staLM \cite{Z_Xie_Learning} & 88.88 & 90.17 & 96.06 & 96.58 \\
		VGG-DBLSTM \cite{K_Chen_Compact} & 87.49 & 87.98 & 97.03 & 97.29 \\
		CharNet-DBLSTM \cite{K_Chen_Compact} & 87.10 & 87.71 & 96.87 & 97.15 \\
		Liu et al. \cite{M_Liu_Distilling} & 91.36 & 92.37 & 94.89 & 95.70 \\
		{\bl Peng et al. \cite{D_Peng_Towards}} & \textbf{\bl 95.05} & \textbf{\bl 95.46} & {\bl \underline{97.36}} & {\bl \underline{97.63}} \\
		\hline 
		\textbf{Ours (tri-gram LM)} & \underline{94.46} & \underline{94.67} & 96.64 & 97.28 \\
		\textbf{\bl Ours (Transformer-based LM)} & {\bl \underline{94.46}} & {\bl \underline{94.67}} & \textbf{\bl 97.89} & \textbf{\bl 98.06} \\
		\hline 
	\end{tabular}
\end{table}

\begin{figure}[b]
	\centering
	\includegraphics[width=\columnwidth]{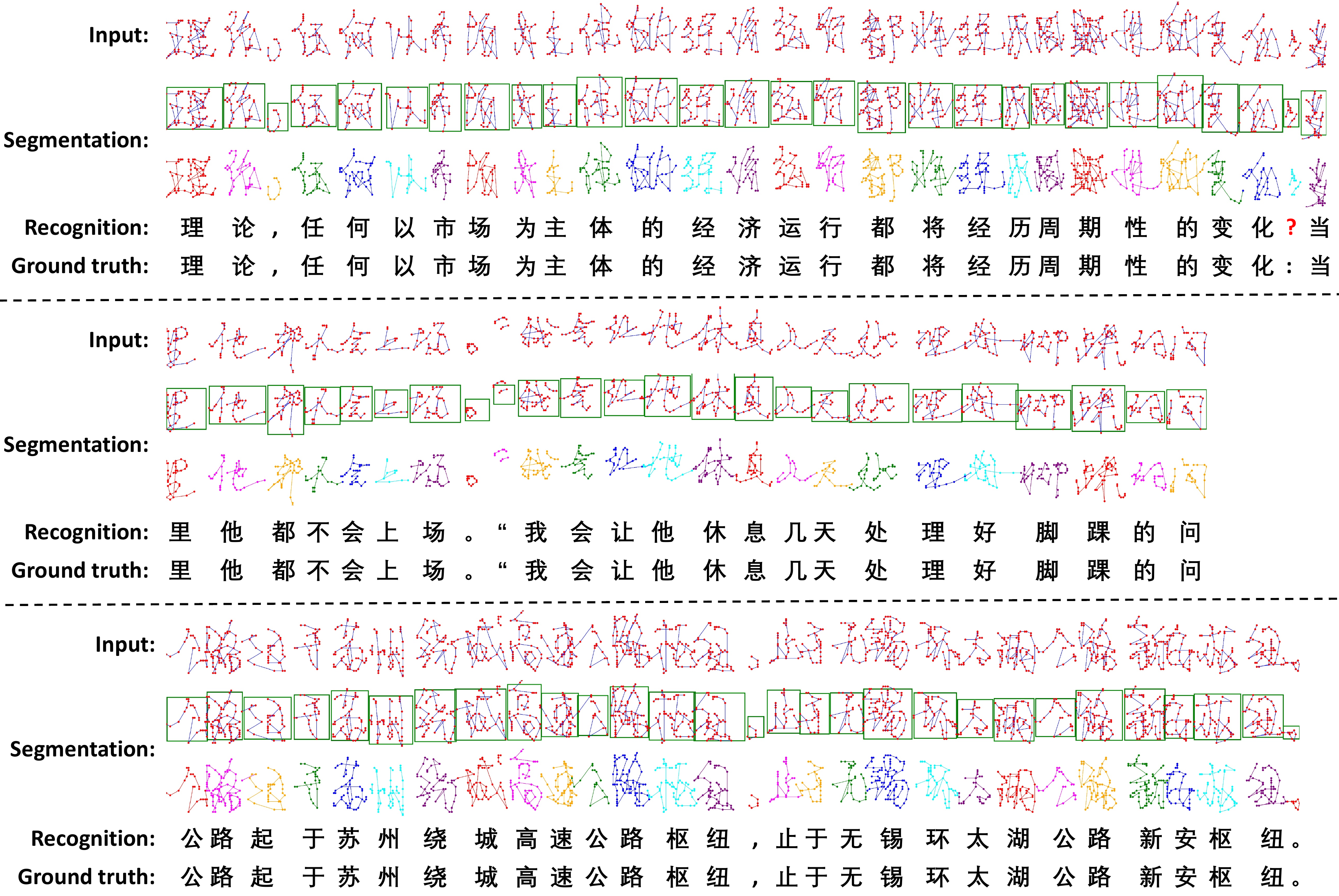}
	\caption{Visualization results of ICDAR2013-Online. The recognition results without language model are presented.}
	\label{fig_exp_vis_icdar2013-online}
\end{figure}

\subsubsection{Experimental Results}
\label{sec_icdaronline_results}

In Table \ref{tbl_exp_ICDAR2013-online}, we compare our methods with existing approaches on ICDAR2013-Online.
{\bl When language models are not used, the performance of our method is comparable to state-of-the-art performance. Although the method proposed by \cite{D_Peng_Towards} performs slightly better, the advantage of our method is that it can produce character segmentation results. When using language models, our approach equipped with the Transformer-based language model outperforms existing methods, including the method in \cite{D_Peng_Towards} which also adopts a Transformer-based language model.}

Fig. \ref{fig_exp_vis_icdar2013-online} illustrates the visualization results of ICDAR2013-Online. According to the predicted bounding boxes, we can 
divide the pen-tip trajectory into several segments, each of which corresponds to one character. Specifically, the points within a bounding box belong to the corresponding character. For the points that are not within any bounding boxes or are within multiple bounding boxes, they belong to the character corresponding to the closest bounding box. As shown in Fig. \ref{fig_exp_vis_icdar2013-online}, both the segmentation and recognition results are very accurate.

\subsection{Experiments on SCUT-HCCDoc}

\begin{table}[b]
	\centering
	\caption{Comparison with existing methods on SCUT-HCCDoc}
	\label{tbl_exp_SCUT-HCCDoc}
	\begin{tabular}{lcc}
		\hline
		Method & AR & CR \\
		\hline
		CTC-based \cite{B_Shi_CRNN} & 87.46 & 88.83 \\
		Attention-based \cite{C_JanK_Attention} & 83.30 & 84.81 \\
		{\bl DAN \cite{T_Wang_Decoupled}} & {\bl 83.53} & {\bl 85.41}\\
		{\bl Liu et al. \cite{liu2021searching}} & {\bl 89.06} & {\bl 90.12} \\
		\hline
		\textbf{Ours} & \textbf{90.71} & \textbf{92.01} \\
		\hline
	\end{tabular}
\end{table}

\begin{table*}[b]
	\centering 
	\caption{Effectiveness of the weakly supervised learning and contextual regularization}
	\label{tbl_exp_ablation}
	\begin{tabular}{lcccccc}
		\hline
		\multirow{2}*{Method} & \multicolumn{2}{c}{ICDAR2013-Online} & \multicolumn{2}{c}{ICDAR2013-Offline} & \multicolumn{2}{c}{SCUT-HCCDoc} \\
		\cmidrule(lr){2-3} \cmidrule(lr){4-5} \cmidrule(lr){6-7}
		& AR & CR & AR & CR & AR & CR \\
		\hline 
		Baseline & 56.18 & 56.90 & 59.68 & 60.38 & 1.14 & 1.25 \\
		+Weakly supervised learning & 93.01 & 93.23 & 93.05 & 93.30 & 90.00 & 91.37 \\
		+Contextual regularization & \textbf{94.46} & \textbf{94.67} & \textbf{94.50} & \textbf{94.76} & \textbf{90.71} & \textbf{92.01} \\
		\hline
	\end{tabular}
\end{table*}

\subsubsection{Experimental Settings}
First, the model is pretrained using the synthetic data (without semantic information) for SCUT-HCCDoc. Then, both the 93,411 real samples from the training set of SCUT-HCCDoc and the synthetic samples (without semantic information) for SCUT-HCCDoc are utilized to train the model. Because the vocabularies of SCUT-HCCDoc and the corpora described in Section \ref{sec_exp_transcription} are very different, the synthetic samples during the training stage are also synthesized without semantic information. Moreover, the ratios of the real and synthetic samples in a batch are 0.7 and 0.3, respectively. After the model training, the 23,218 samples from the testing set of SCUT-HCCDoc are adopted to evaluate the method. The number of character categories is set to 6,109.

\subsubsection{Data Preprocessing} Both the real and synthetic text-line images are resized to a height of 128 pixels while maintaining their aspect ratios.

\subsubsection{Experimental Results}
\label{sec_exp_hccdoc_results}

In Table \ref{tbl_exp_SCUT-HCCDoc}, we compare our method with {\bl existing approaches on SCUT-HCCDoc.} Specifically, the latest results of the CTC/attention-based approaches \cite{B_Shi_CRNN,C_JanK_Attention}, which were updated by the authors of \cite{H_Zhang_SCUT-HCCDoc} at their website\footnote[2]{https://github.com/HCIILAB/SCUT-HCCDoc\_Dataset\_Release\label{hccdoc_web}}, are presented in Table \ref{tbl_exp_SCUT-HCCDoc}. {\bl The performances of the other two methods \cite{liu2021searching,T_Wang_Decoupled} are obtained from our reimplementation based on their official codes\footnote[3]{https://github.com/intel/handwritten-chinese-ocr-samples}\textsuperscript{,}\footnote[4]{https://github.com/Wang-Tianwei/Decoupled-attention-network}.}
It can be observed that our method achieves state-of-the-art performance on SCUT-HCCDoc with an AR of 90.71\% and a CR of 92.01\%. 

\begin{figure}[t]
	\centering
	\vspace{0.5em}
	\includegraphics[width=\columnwidth]{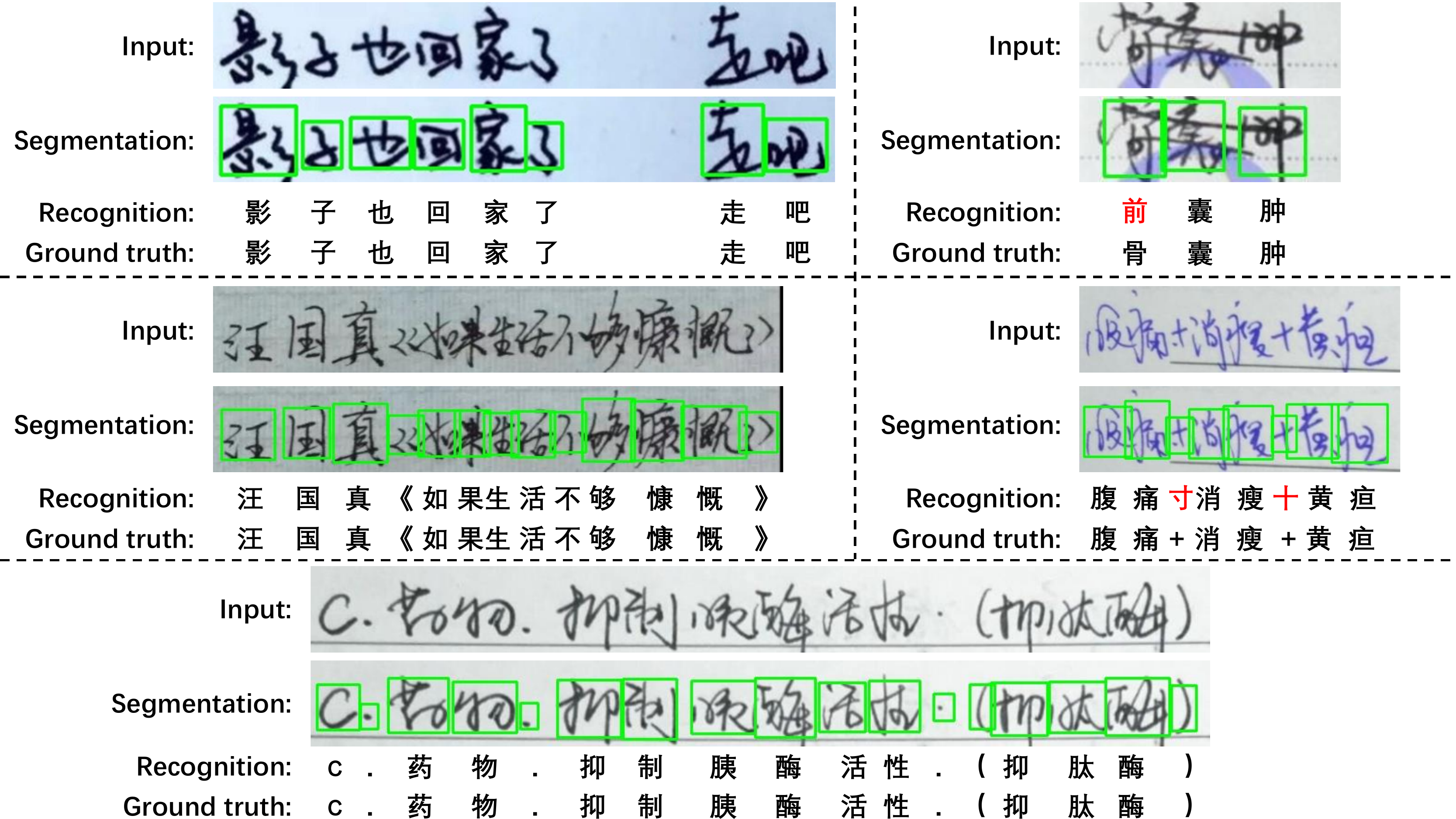}
	\caption{Visualization results of SCUT-HCCDoc.}
	\label{fig_exp_HCCDoc}
\end{figure}

The visualization results of SCUT-HCCDoc are shown in Fig. \ref{fig_exp_HCCDoc}. Although the synthetic samples for SCUT-HCCDoc are composed of simple characters generated from font files and white backgrounds as illustrated in Fig. \ref{Fig_Exp_Synth}(c), our method can still make full use of them and learn to segment and recognize characters of complex real samples. From the visualizations in Fig. \ref{fig_exp_HCCDoc}, we can observe that our method can handle various writing styles including illegible handwriting and is unaffected by the noises such as illumination and the interference from the background. Especially for the sample at the top-right of Fig. \ref{fig_exp_HCCDoc}, where the text is crossed out, the characters can still be successfully segmented and recognized, which verifies the robustness of our approach.

\subsection{Ablation Studies}

In Table \ref{tbl_exp_ablation}, we conduct ablation analysis to demonstrate the effectiveness of each component of our method. 

The baseline method represents the pretrained model that uses only synthetic data. It can be seen that training the network using only synthetic data leads to very poor performance. Especially for SCUT-HCCDoc where the real images are very different from the synthetic samples, the baseline can only achieve an extremely low accuracy with an AR of 1.14\% and a CR of 1.25\%. 

When the proposed weakly supervised learning method is adopted, real samples with only {\bl transcript} annotations can also be utilized to train the model. Taking advantage of the effective design of the weakly supervised learning method, the model can make full use of the general knowledge acquired from simple synthetic data and be adapted to handle the real sample by learning from its past successful predictions. Table \ref{tbl_exp_ablation} shows that performance can be significantly improved. Even though the pretrained model for SCUT-HCCDoc has extremely poor performance, our weakly supervised learning method can also work very well.

ConR is aimed at integrating contextual information into the feature maps for character classification.
The results in Table \ref{tbl_exp_ablation} demonstrate that ConR can bring remarkable improvement.

\begin{table*}[t]
	\centering 
	\caption{Effectiveness of the weakly supervised learning method.}
	\label{tbl_exp_wsl}
	\begin{tabular}{lcccccc}
		\hline
		\multirow{2}*{Method} & \multicolumn{2}{c}{ICDAR2013-Online} & \multicolumn{2}{c}{ICDAR2013-Offline} & \multicolumn{2}{c}{SCUT-HCCDoc} \\
		\cmidrule(lr){2-3} \cmidrule(lr){4-5} \cmidrule(lr){6-7}
		& AR & CR & AR & CR & AR & CR \\
		\hline 
		\textit{Text length}-based \cite{L_Xing_Convolutional,Y_Baek_Character} & 51.45 & 51.54 & 63.89 & 64.94 & 86.98 & 90.13\\
		\textbf{Ours} & \textbf{93.01} & \textbf{93.23} & \textbf{93.05} & \textbf{93.30} & \textbf{90.00} & \textbf{91.37} \\
		\hline
	\end{tabular}
\end{table*}

\subsection{Effectiveness of Weakly Supervised Learning}
\label{sec_effect_WSL}

We further analyze the effectiveness of the proposed weakly supervised learning method. In our method, we distinguish the correct prediction by computing the edit distance between the recognition result and {\bl transcript} annotation. Moreover, the pseudo bounding boxes are updated in a weighted-sum manner, considering all the correct predictions in previous iterations and their scores. 

However, some existing approaches \cite{L_Xing_Convolutional,Y_Baek_Character} that involve weakly supervised learning follow a very simple pipeline. If the lengths of the recognition result and {\bl transcript} annotation are equal, the predicted bounding boxes are directly viewed as the ground truth. Following their ideas, we replace our weakly supervised learning method with a new one named \textit{Text Length}, where the pseudo bounding boxes $B_{pse}$ are updated as
\begin{equation}
	\small
	B_{pse} = 
	\begin{cases}
		R_{seg}, & l_{pr} = l_{gt}, \\
		B_{pse}, & otherwise, \\
	\end{cases}
\end{equation}
where $R_{seg}$, $l_{pr}$, and $l_{gt}$ are the segmentation result, the length of the recognition result, and the length of the {\bl transcript} annotation, respectively (as specified in Section \ref{sec_methodology_wsl}).

As presented in Table \ref{tbl_exp_wsl}, our method outperforms the \textit{Text length} by a large margin, especially for ICDAR2013-Online and ICDAR2013-Offline that contain long texts. All models in Table \ref{tbl_exp_wsl} are trained without contextual regularization. The criterion of \textit{Text length} (i.e., the length of predictions and annotations are equal) cannot guarantee the quality of the predictions used for updating the pseudo bounding boxes, and is rarely satisfied for long texts and pretrained models with low accuracy. Moreover, the pseudo bounding boxes are updated by copying new predictions, which may make the training easily be interfered with by poor predictions. However, our method is based on the observation that correctly recognized characters are likely to have accurate bounding boxes. Thus only the characters that are matched as ``equal'' in computing edit distance are selected to ensure the quality of bounding boxes. Furthermore, the updating of pseudo bounding boxes is also carefully designed to suppress the impact of potential poor predictions.

\begin{table}[t]
	\centering 
	\caption{Comparison with fully supervised learning on offline subset of ICDAR2013 competition dataset}
	\label{tbl_exp_fully}
	\begin{tabular}{lcc}
		\hline 
		Method & AR & CR \\
		\hline
		Fully supervised & 94.43 & 94.64 \\
		Weakly supervised & \textbf{94.50} & \textbf{94.76} \\
		\hline
	\end{tabular}
\end{table}

By directly using the character bounding boxes provided by the annotations of CASIA-HWDB2.0-2.2, we can also train our model under full supervision. In Table \ref{tbl_exp_fully}, we compare the performance of our method on ICDAR2013-Offline under different supervisions{\bl, where the two models are trained with the same experimental settings as specified in Sections \ref{sec_implement_detail} and \ref{sec_exp_icdaroffline}}. It can be seen that the weakly supervised model can even reach slightly higher AR and CR compared with the fully supervised counterpart, which verifies the effectiveness of our weakly supervised learning method. {\bl The superior performance under weak supervision may be due to the iterative pseudo bounding box updating mechanism. Specifically, the model is supervised by different pseudo bounding boxes every time perceiving the same text-line image, which may play a role of regularization.}

\begin{table}[t]
	\centering
	\caption{Comparison of the performance with and without the BLSTM layers. The ``Ratio'' column presents the ratio of the time consumed on the BLSTM layers to the time consumed on the entire network.}
	\label{tbl_exp_conr}
	\begin{tabular}{lccccc}
		\hline
		\multirow{2}*{Dataset} & \multicolumn{2}{c}{Without BLSTM} & \multicolumn{3}{c}{With BLSTM} \\
		\cmidrule(lr){2-3} \cmidrule(lr){4-6}
		& AR & CR & AR & CR & Ratio \\
		\hline
		ICDAR2013-Online & \textbf{94.46} & \textbf{94.67} & 94.38 & 94.62 & 42\% \\
		ICDAR2013-Offline & \textbf{94.50} & \textbf{94.76} & 94.46 & 94.72 & 42\% \\
		SCUT-HCCDoc & 90.71 & 92.01 & \textbf{90.85} & \textbf{92.11} & 23\% \\
		\hline
	\end{tabular}
\end{table}

\subsection{Effectiveness of Contextual Regularization}
\label{sec_exp_effect_ConR}

In Table \ref{tbl_exp_conr}, experiments are conducted for further analysis of ConR. 

First, we investigate the performance of the model if the BLSTM layers and the subsequent character classification results are adopted during inference, instead of using the character classification results from feature map $f_{cls}$. As shown in Table \ref{tbl_exp_conr}, the performances with and without the BLSTM layers are nearly the same. Specifically, compared with the counterpart with BLSTM layers, the AR and CR without BLSTM layers can be slightly higher on ICDAR2013-Online and ICDAR2013-Offline, and only drop a little on SCUT-HCCDoc. Based on the above results, we can conclude that ConR can really integrate the contextual information into the feature $f_{cls}$ before BLSTM layers.

{\bl As for the inference speed, the two BLSTM layers can consume more than 40\% of the inference time of the entire network if they are used for context modeling, as demonstrated by the ``Ratio'' column of Table \ref{tbl_exp_conr}.} This is attributed to the non-parallel running of the BLSTM layer and the long texts that are common in Chinese documents. With the help of ConR, the recurrent layers are not required during inference, thus significantly improving the speed.

\begin{table}[t] 
	\centering 
	\caption{Comparison with CTC-based and attention-based methods}
	\label{tbl_exp_ctc_attention}
	\begin{tabular}{llccc}
		\hline 
		Dataset & Method & AR & CR & Speed \\
		\hline 
		\multirow{2}*{ICDAR2013-Online} & CTC-based \cite{M_Liu_Distilling} & 91.36 & 92.37 & 62fps\\
		& Attention-based \cite{B_Shi_Robust} & 85.35 & 85.84 & 16fps\\
		& \textbf{Ours} & \textbf{94.46} & \textbf{94.67} & \textbf{70fps}\\
		\hline
		\multirow{3}*{ICDAR2013-Offline} & CTC-based \cite{C_Xie_High} & 91.55 & 92.13 & 64fps \\
		& Attention-based \cite{B_Shi_Robust} & 84.79 & 85.90 & 16fps \\
		& \textbf{Ours} & \textbf{94.50} & \textbf{94.76} & \textbf{70fps} \\
		\hline
		\multirow{3}*{SCUT-HCCDoc} & CTC-based \cite{B_Shi_CRNN} & 87.64 & 88.83 & 75fps \\
		& Attention-based \cite{C_JanK_Attention} & 83.30 & 84.81 & 52fps \\
		& \textbf{Ours} & \textbf{90.71} & \textbf{92.01} & \textbf{97fps} \\
		\hline
	\end{tabular}
\end{table}

\subsection{Comparison with CTC and Attention}
\label{sec_exp_cmp_ctc_attn}
Table \ref{tbl_exp_ctc_attention} compares our method with existing widely used CTC/attention-based approaches.

For ICDAR2013-Online and ICDAR2013-Offline, two representative CTC-based methods \cite{M_Liu_Distilling,C_Xie_High} that follow mainstream CNN+RNN+CTC or RNN+CTC architectures are adopted for comparison. Owing to the lack of attention-based methods in previous literature, we reimplement the sequence recognition network of \cite{B_Shi_Robust} on ICDAR2013-Online and ICDAR2013-Offline with the same training data and preprocessing as ours. Because the network \cite{B_Shi_Robust} is designed for offline images, the path signature feature maps (Section \ref{sec_exp_ps}) are also adopted for online texts. For SCUT-HCCDoc, we compare our method with CTC/attention-based approaches \cite{B_Shi_CRNN,C_JanK_Attention} adopted in \cite{H_Zhang_SCUT-HCCDoc} (same as Table \ref{tbl_exp_SCUT-HCCDoc}). Furthermore, we also test the inference speed of these methods and ours using an NVIDIA GTX 1080ti GPU with 11GB of memory. 

The results in Table \ref{tbl_exp_ctc_attention} show that our method outperforms existing CTC/attention-based approaches in terms of AR and CR. {\bl Moreover, owing to the parallel computing characteristic of the fully convolutional architecture, our method exhibits a higher inference speed compared with the CTC/attention-based methods that use recurrent layers for contextual modeling. 
} 
In addition, our method can produce character segmentation results, whereas CTC/attention-based approaches cannot.

\subsection{Reading Chinese Text in the Wild}
\label{sec_rects}

In this section, we further explore the potential of our method to be extended to the recognition of  Chinese text in the wild. The experiments are conducted using ReCTS-25k \cite{R_Zhang_ReCTS}, which contains 25,000 signboard images (20,000 for training and 5,000 for testing). {\bl There are 108,963 and 10,789 text lines cropped from the training and testing sets, respectively. We further divide the text lines from the training set into 90,763 samples for training and 18,200 samples for validating. } The synthetic data is synthesized in the same way as in Section \ref{sec_data_synthesis}, using white background and characters from font files. The number of character categories is 4,134. Other details follow the experiments on SCUT-HCCDoc and Section \ref{sec_implement_detail}.

\begin{table}[t]
	\centering 
	\begin{threeparttable}
		\caption{Performance on ReCTS-25k dataset}
		\label{tbl_exp_rects}
		\begin{tabular}{lccc}
			\hline 
			Method & {\bl Validating NED} & {\bl Testing NED} & Speed \\
			\hline
			CTC-based \cite{B_Shi_CRNN} & 80.21 & {\bl 83.40} & 86fps \\
			Attention-based \cite{B_Shi_Robust} & 82.02 & {\bl 86.46}  & 69fps \\
			{\bl SANHL\_v1\tnote{*} \cite{R_Zhang_ReCTS}} & - & \textbf{\bl 95.55} & - \\
			\hline 
			\textbf{Ours} & \textbf{86.66} & {\bl 90.70} & \textbf{108fps} \\
			\hline	
		\end{tabular}
		\begin{tablenotes}
			{\bl \item[*] The winner of the ICDAR 2019 ReCTS competition \cite{R_Zhang_ReCTS}.}
		\end{tablenotes}
	\end{threeparttable}
\end{table}

\begin{figure}[t]
	\centering
	\vspace{0.5em}
	\includegraphics[width=\columnwidth]{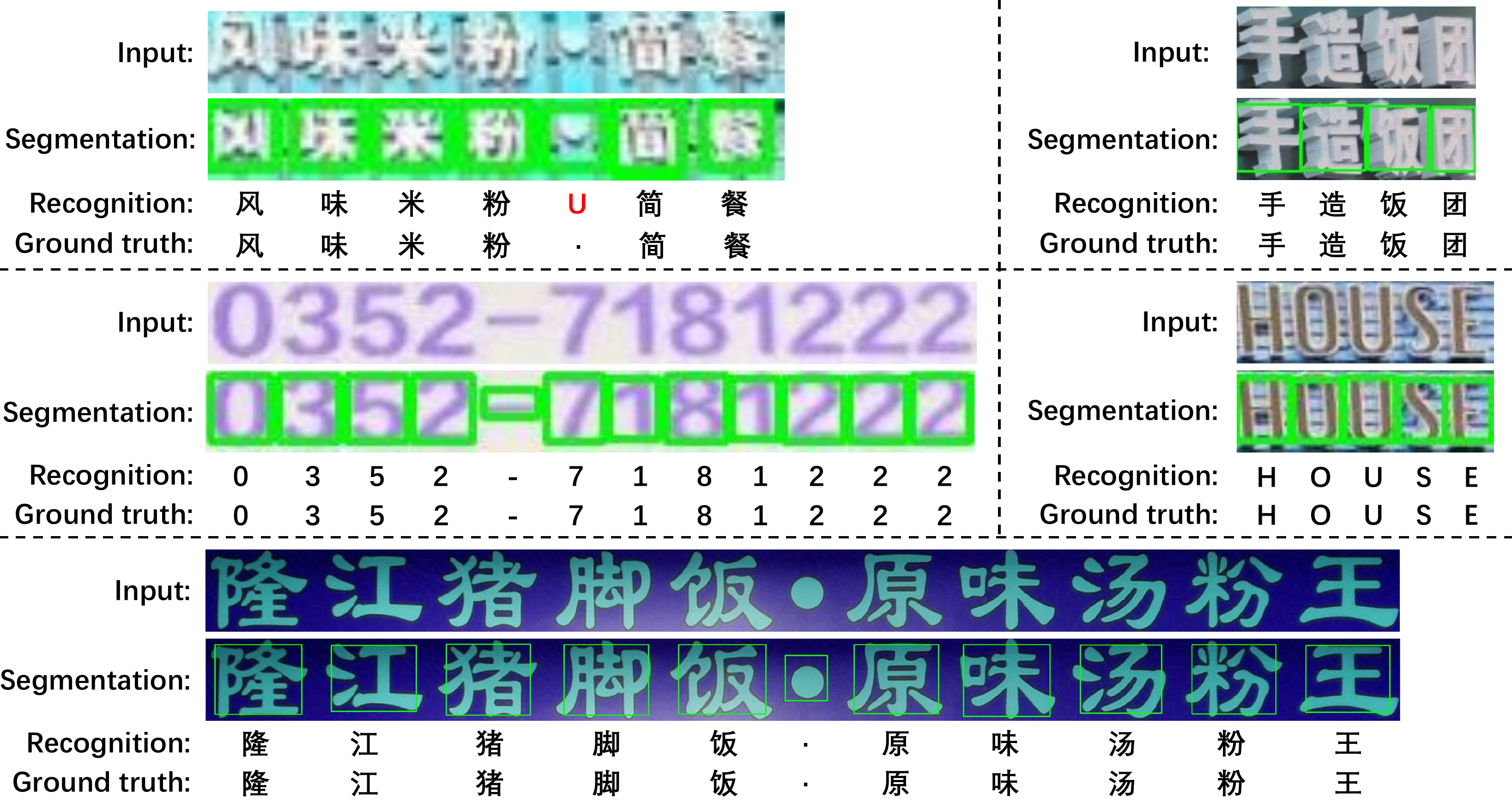}
	\caption{Visualization results of ReCTS-25k.}
	\label{fig_exp_vis_rects}
\end{figure}

As shown in Table \ref{tbl_exp_rects}, our method achieves a normalized edit distance (NED) \cite{R_Zhang_ReCTS} of 86.66\% {\bl on the validating set and 90.70\% on the testing set.} Some visualization results are shown in Fig. \ref{fig_exp_vis_rects}. 
When tested using an NVIDIA GTX 1080ti GPU, our method performs better with a higher inference speed. 
{\bl However, our approach is inferior to the winning method (SNAHL\_v1 with 95.55\% NED) of the competition \cite{R_Zhang_ReCTS}, which uses an ensemble of three types of models and a large amount of external data.}
To conclude, our method has the potential of being extended to other applications besides online and offline HCTR. 

\begin{figure}[b]
	\centering
	\includegraphics[width=0.75\columnwidth]{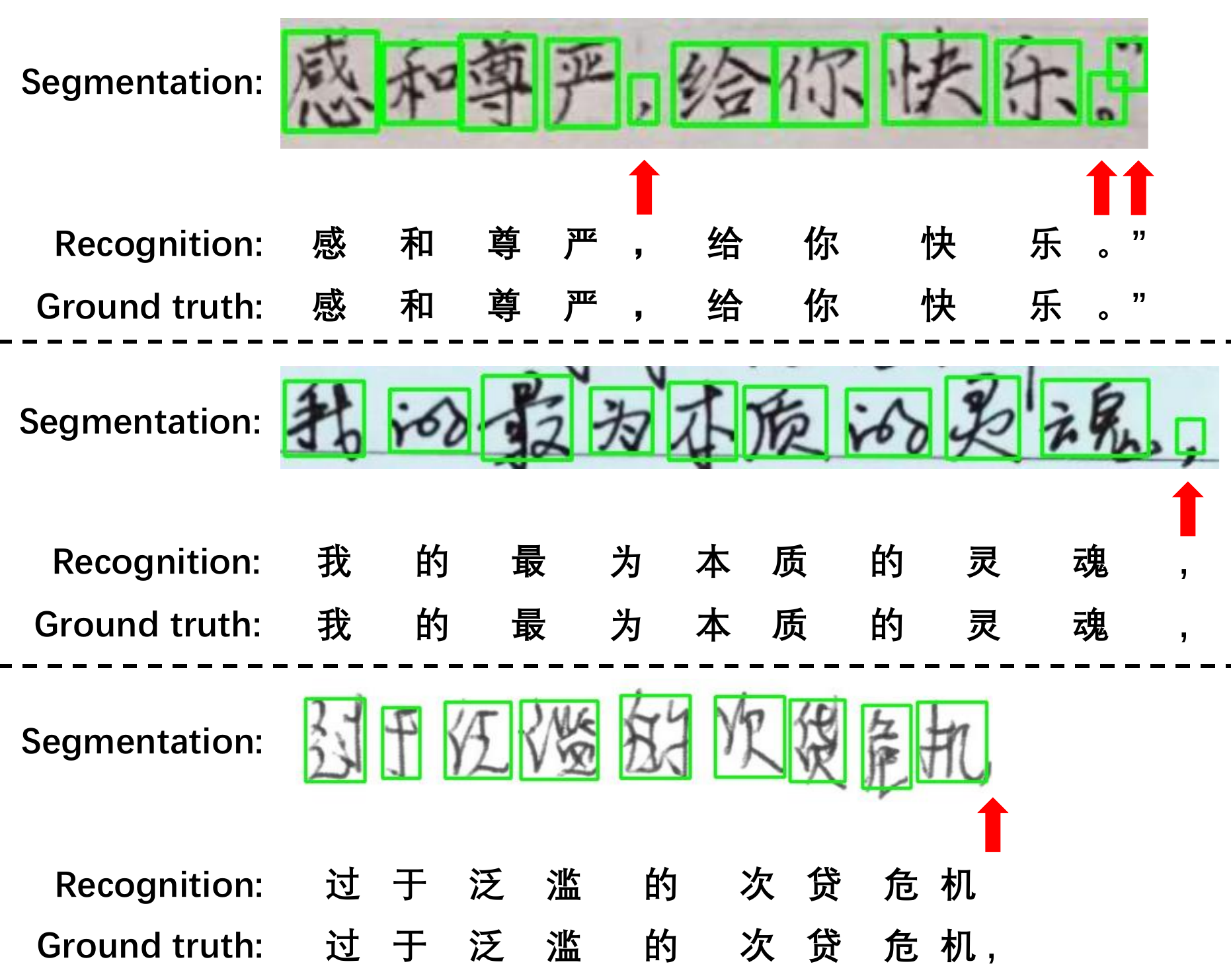}
	\caption{Failure cases of the segmentation of punctuations.}
	\label{fig_exp_error_ana}
\end{figure}

\subsection{Error Analysis}
The weakness of our method lies in the segmentation of punctuations. Fig. \ref{fig_exp_error_ana} presents the failure cases of punctuation segmentation (indicated by red arrows). Compared with Chinese characters, punctuations are much smaller and have more flexible locations that could be the top or bottom of the text line. However, punctuations are overwhelmed by Chinese characters in the training data, which may be the major cause of the poor segmentation of punctuations. Nevertheless, owing to the decoupled design of character segmentation and classification, the recognition results can still be correctly predicted in the first two cases of Fig. \ref{fig_exp_error_ana}.
Moreover, the punctuation could be very close to its previous character especially for handwritten texts, which makes it difficult to distinguish. For example, the comma in the third case of Fig. \ref{fig_exp_error_ana} is omitted by our method.

\begin{figure}[t]
	\centering 
	\includegraphics[width=0.6\columnwidth]{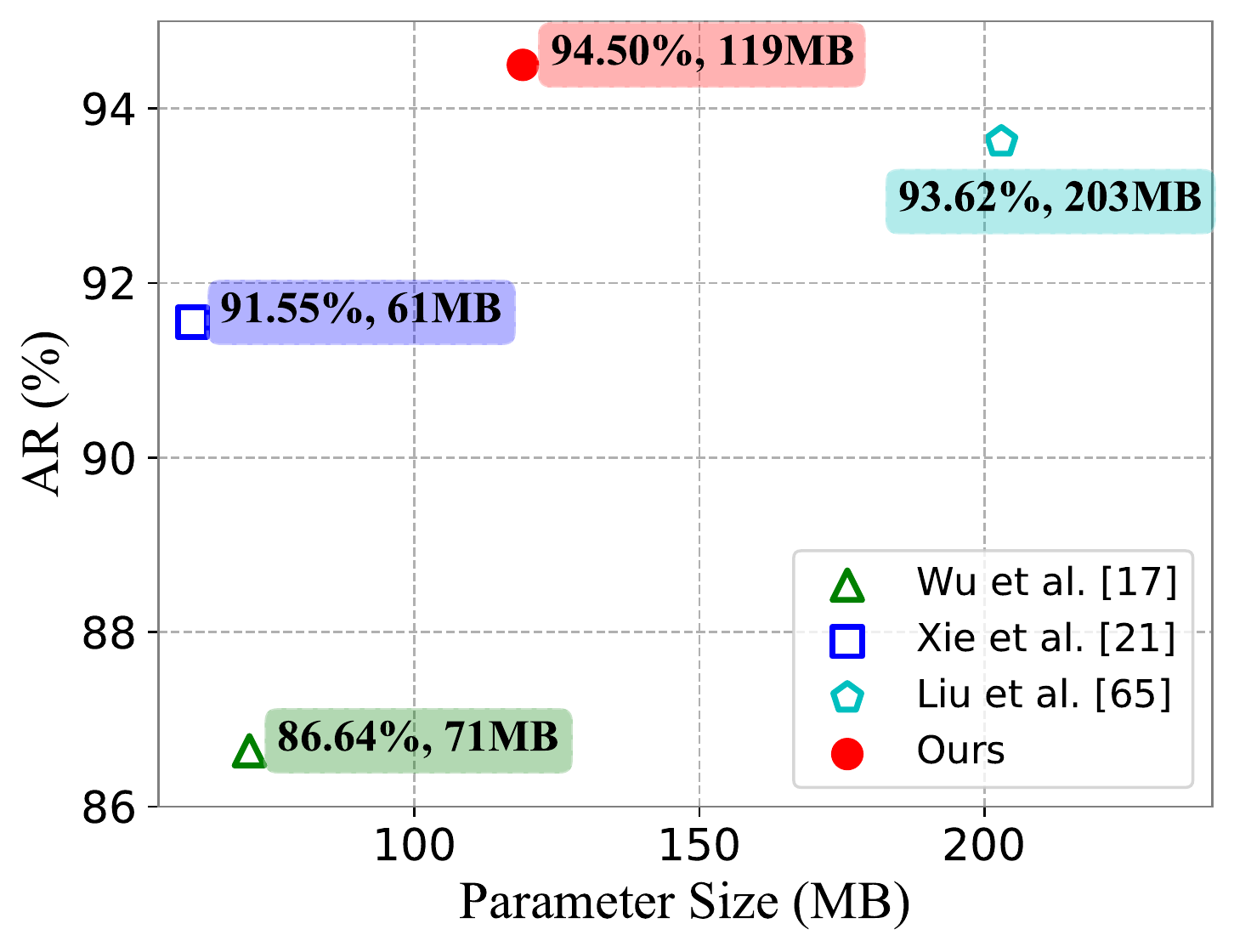}
	\caption{\bl AR versus parameter size on ICDAR2013-Offline.}
	\label{fig_exp_ar_parametersize}
\end{figure}

{\bl \subsection{Discussion}
\label{sec_exp_discussion}
\subsubsection{Parameter Size}
\label{sec_exp_discuss_parametersize}

Our method is basically for offline text recognition because online texts are also expressed as offline representations. Thus, we illustrate the AR versus parameter size of existing methods on ICDAR2013-Offline in Fig. \ref{fig_exp_ar_parametersize}. Note that only methods whose parameter sizes were reported in previous literature are presented. It can be observed that our model has a good trade-off between AR and parameter size. Especially compared with the method in \cite{liu2021searching}, our method achieves better performance with approximately half of the parameters.}

{\bl \subsubsection{Training Time}
\label{sec_exp_discuss_trainingtime}
As described in Section \ref{sec_implement_detail}, there are two stages of our model training, i.e., the pretraining and training stages. Using an NVIDIA RTX 2080ti GPU, it takes approximately 125 hours in total to train a model for ICDAR2013-Offline. Specifically, the pretraining and training stages take 14 hours and 111 hours, respectively. The time required to train a model for ICDAR2013-Online is nearly the same because we translate the online pen-tip trajectories into image-like representations. In previous literature, there are few studies which reported their training time.
For example, the method in \cite{liu2021searching} for ICDAR2013-Offline requires approximately 80 hours, and the method in \cite{M_Liu_Distilling} for ICDAR2013-Online requires 102 hours. Compared with these methods, the training time of our method is slightly longer but still acceptable. Because most procedures of the weakly supervised learning run on CPU in our implementation, the training speed could be accelerated by migrating them to GPU in the future.
}

{\bl \subsubsection{Failure Case of Weakly Supervised Learning} 
\label{sec_exp_discuss_failurecase_wsl}
If the ground truth consists of multiple same characters, such as ``AAA'', but only ``AA'' is recognized, we can not determine which two ``A''s in the ground truth are correctly recognized by calculating edit distance. This issue could lead to inaccurate pseudo bounding boxes because they may be updated using unsuitable predictions. However, such a situation rarely occurs in real data that contains natural texts. Thus, the model training is almost unaffected. Nevertheless, this problem can be explored by incorporating spatial constraints or feature similarities in the future.}

\section{Conclusion}
\label{sec_conclusion}
In this paper, we propose a novel segmentation-based method for online and offline HCTR. In contrast to previous oversegmentation-based approaches, we formulate a brand-new segmentation-based text recognition framework that end-to-end segments and recognizes characters through fully convolutional networks with high efficiency and accuracy. To address the high cost of character segmentation annotations, a new weakly supervised learning method is proposed to enable the network to be trained using only {\bl transcript} annotations. Owing to the absence of context modeling in the fully convolutional architecture, we design a contextual regularization method to integrate contextual information into extracted features without affecting the inference speed. Extensive experiments on CASIA-HWDB, CASIA-OLHWDB, ICDAR2013, and SCUT-HCCDoc, demonstrate the superiority of our method over existing approaches. To the best of our knowledge, our method may be the first to achieve state-of-the-art performance on both online and offline HCTR. An additional trial on ReCTS-25k demonstrates the potential of our method out of the field of HCTR. We hope this work will spark further research beyond the realms of prevalent CTC/attention-based methods.

\section*{Acknowledgement}
This research is supported in part by NSFC (Grant No.: 61936003, 61771199), and GD-NSF (no. 2017A030312006).

\ifCLASSOPTIONcaptionsoff
  \newpage
\fi

\begin{IEEEbiography}[{\includegraphics[width=1in,height=1.25in,clip,keepaspectratio]{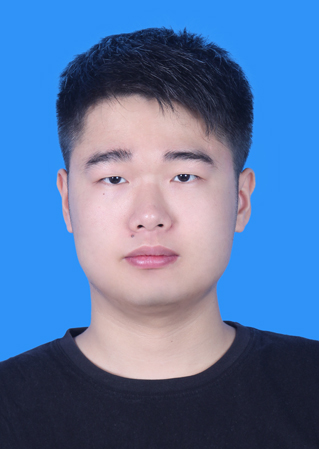}}]{Dezhi Peng} received the B.S. degree in information engineering from South China University of Technology in 2019. He is currently pursuing the Ph.D. degree in information and communication engineering at South China University of Technology. His research interests include optical character recognition, document analysis and recognition, and handwriting text recognition.
\end{IEEEbiography}
\begin{IEEEbiography}[{\includegraphics[width=1in,height=1.25in,clip,keepaspectratio]{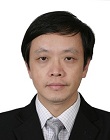}}]{Lianwen Jin} received the B.S. degree from the University of Science and Technology of China, Anhui, China, and the Ph.D. degree from the South China University of Technology, Guangzhou, China, in 1991 and 1996, respectively. He is currently a Professor with the School of Electronic and Information Engineering, South China University of Technology. He is the author of more than 100 scientific papers. Dr. Jin was a recipient of the award of New Century Excellent Talent Program of MOE in 2006 and the Guangdong Pearl River Distinguished Professor Award in 2011. His research interests include computer vision, optical character recognition, handwriting analysis and recognition, machine learning, deep learning, and intelligent systems.
\end{IEEEbiography}
\begin{IEEEbiography}[{\includegraphics[width=1in,height=1.25in,clip,keepaspectratio]{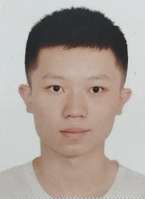}}]{Weihong Ma} received the B.S. degree from the school of Electronic and Information Engineering at the South China University of Technology, Guangzhou, China in 2019. He is currently pursuing the master degree in information and communication engineering at the South China University of Technology, Guangzhou, China. His current research interests include deep learning, scene text detection, and document analysis.
\end{IEEEbiography}
\begin{IEEEbiography}[{\includegraphics[width=1in,height=1.25in,clip,keepaspectratio]{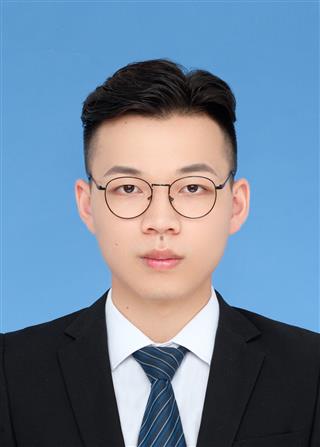}}]{Canyu Xie} received the B.S. degree in electronic science and technology at the South China University of Technology, Guangzhou, China in 2019. He is currently pursuing the master degree in electronic and communication engineering at South China University of Technology, Guangzhou, China. His current research interests include computer vision, model compression, and acceleration.
\end{IEEEbiography}
\begin{IEEEbiography}[{\includegraphics[width=1in,height=1.25in,clip,keepaspectratio]{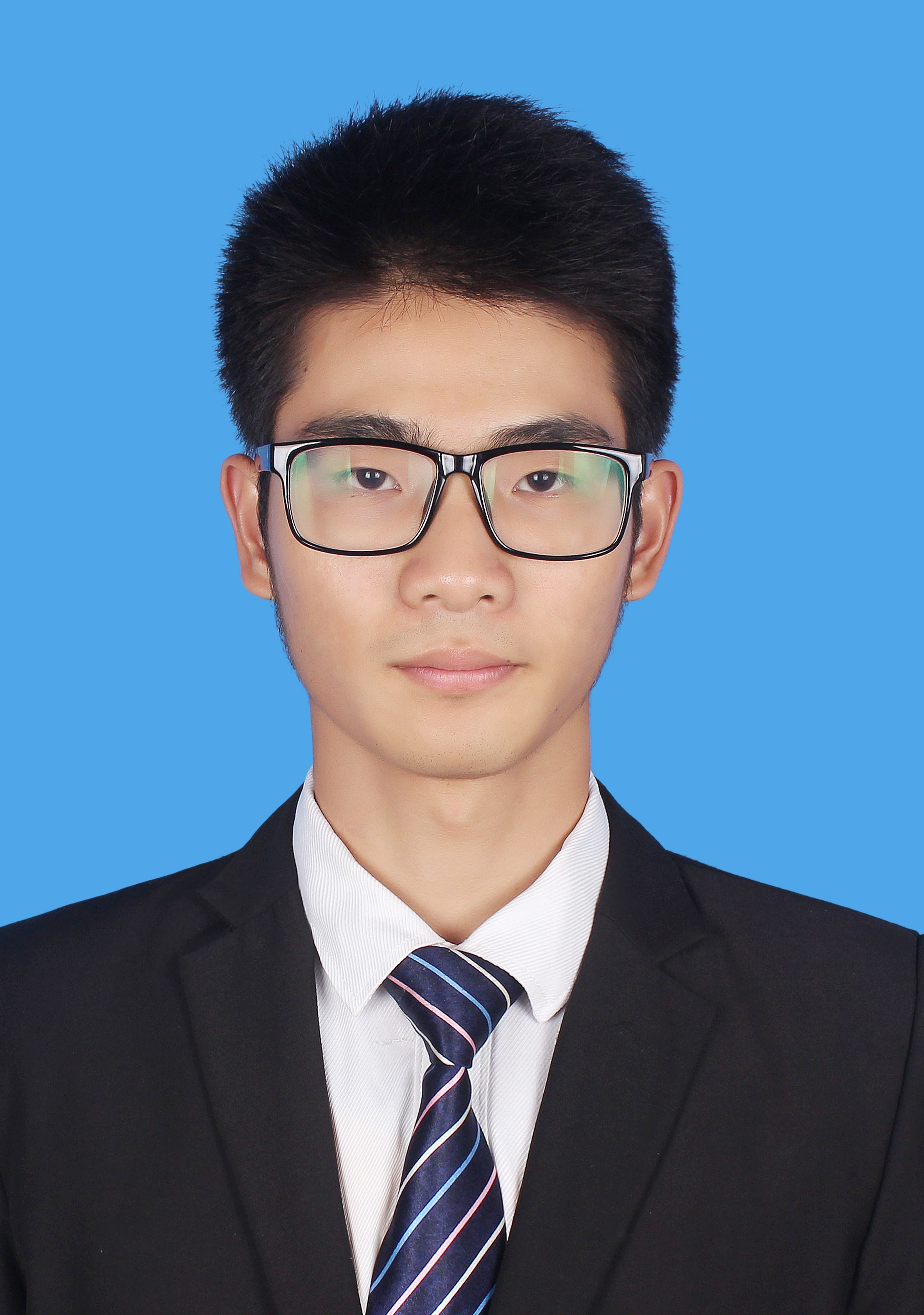}}]{Hesuo Zhang} received the B.S. degree from the school of Electronic and Information Engineering at South China University of Technology, Guangzhou, China in 2019. He is currently pursuing the master degree in information and communication engineering at South China University of Technology, Guangzhou, China. His current research interests include machine learning, deep learning, and handwritten text segmentation and recognition.
\end{IEEEbiography}
\begin{IEEEbiography}
[{\includegraphics[width=1in,height=1.25in,clip,keepaspectratio]{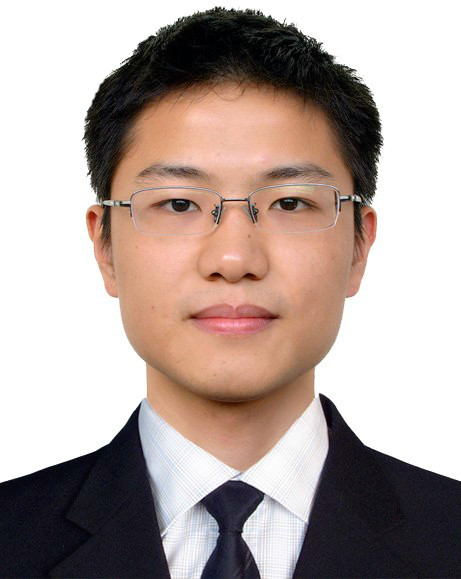}}]{Shenggao Zhu} received the Ph.D. degree in Computer Science from National University of Singapore (NUS), 2017. He got his bachelor in Electronic Engineering and Information Science from University of Science and Technology of China (USTC), 2011. He joined Huawei Cloud in 2017 and now is a Technical Expert. His research interests include computer vision and AI applications.
\end{IEEEbiography}
\begin{IEEEbiography}
[{\includegraphics[width=1in,height=1.25in,clip,keepaspectratio]{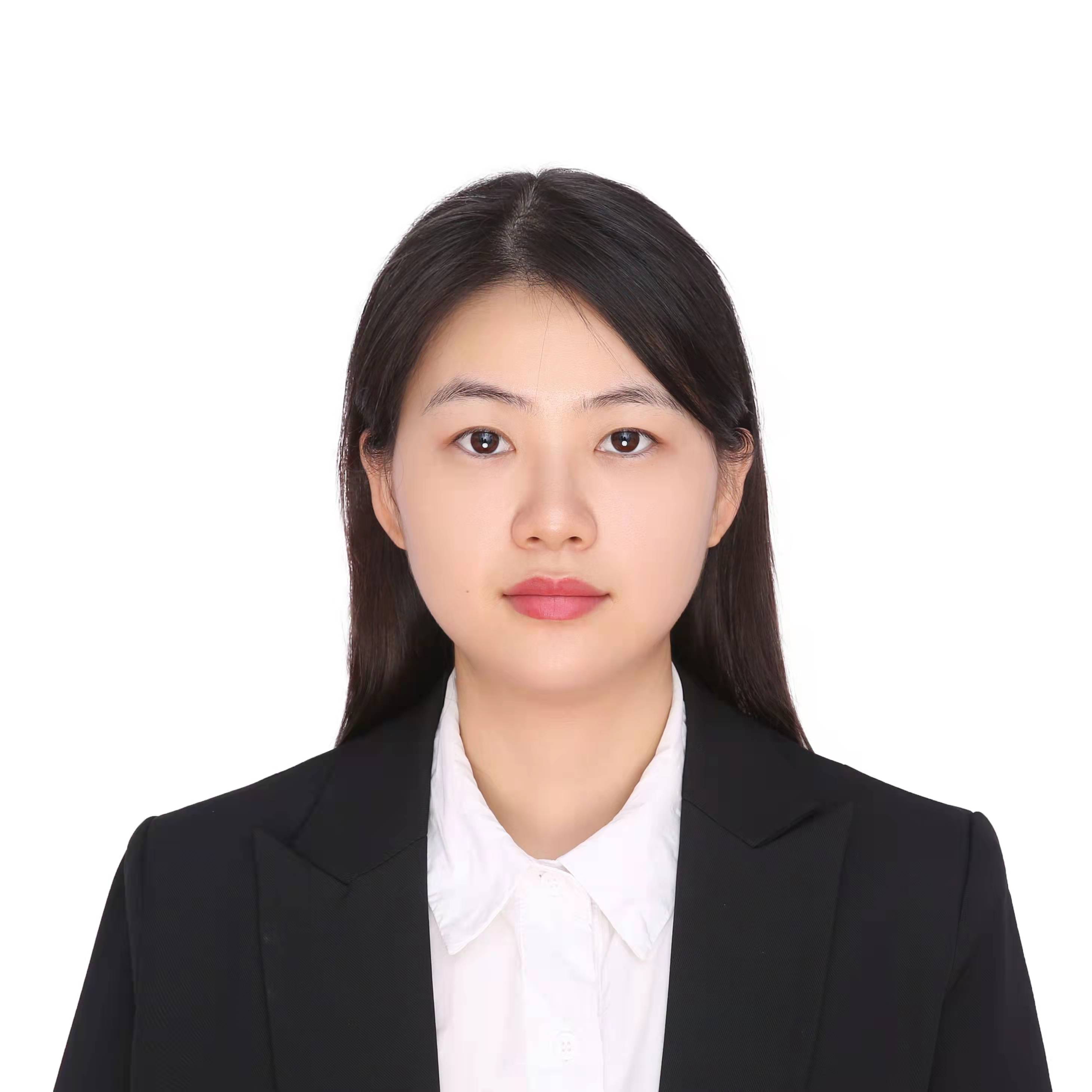}}]{Jing Li} received the B.S. degree from the University of Science and Technology of China, and the Ph.D. degree from the National University of Singapore, in 2013 and 2019, respectively. She is now an AI engineer in Huawei Cloud Computing Technologies. Her research interests include face recognition, optical character recognition and image generation. 
\end{IEEEbiography}
\vfill
\end{document}